\documentclass{article}
\usepackage{iclr,times}
\usepackage[T1]{fontenc}
\usepackage{amsmath,amssymb,booktabs,nicematrix,graphicx,xcolor}
\usepackage{hyperref}
\usepackage{url}
\usepackage[position=bottom,strut=off,hypcap=false]{caption}
\usepackage{placeins}
\usepackage{wrapfig}
\usepackage{needspace}
\newcommand{\AW}{AssemblyWorld}
\newcommand{\AB}{AssemblyWorldBench}

\AddToHook{env/figure/begin}{  \setlength{\abovecaptionskip}{3pt}  \setlength{\belowcaptionskip}{0pt}}
\AddToHook{env/wrapfigure/begin}{  \setlength{\abovecaptionskip}{3pt}  \setlength{\belowcaptionskip}{0pt}}
\AddToHook{env/table/begin}{  \setlength{\abovecaptionskip}{0pt}  \setlength{\belowcaptionskip}{3pt}}
\AddToHook{env/wraptable/begin}{  \setlength{\abovecaptionskip}{0pt}  \setlength{\belowcaptionskip}{3pt}}
\hypersetup{hidelinks,pdfauthor={Jiahao Zhang, Yeying Fan, Moitreya Chatterjee, Suhas Lohit, Bernhard Egger, Tim K. Marks, Anoop Cherian, Stephen Gould},pdftitle={AssemblyWorld: Rethinking 3D Assembly with General-Purpose Agents}}
\newcommand{\SE}{\mathrm{SE}(3)}
\newcommand{\venue}[2]{#1\textsubscript{\scriptsize #2}}

\title{AssemblyWorld: Rethinking 3D Assembly with General-Purpose Agents}
\iclrfinalcopy
\author{Jiahao Zhang$^{1}$\thanks{Equal contribution.}, Yeying Fan$^{3*}$, Moitreya Chatterjee$^{2}$, Suhas Lohit$^{2}$\\
\bfseries Bernhard Egger$^{4}$, Tim K. Marks$^{2}$, Anoop Cherian$^{2}$, Stephen Gould$^{1}$\\[0.5em]
\normalfont\small $^{1}$The Australian National University \quad $^{2}$Mitsubishi Electric Research Laboratories (MERL)\\
\normalfont\small $^{3}$Tsinghua University \quad $^{4}$Friedrich-Alexander-Universit\"at Erlangen-N\"urnberg\\[0.3em]
\normalfont\small \href{https://assemblyworld.github.io/}{\texttt{https://assemblyworld.github.io}}
}
\date{}
\begin{document}
\maketitle
\lhead{}
\pagestyle{plain}
\thispagestyle{plain}
\par\vspace{-20pt}\noindent
\begin{minipage}{\linewidth}
\centering
\setlength{\abovecaptionskip}{3pt}
\setlength{\belowcaptionskip}{0pt}
\includegraphics[width=0.94\linewidth]{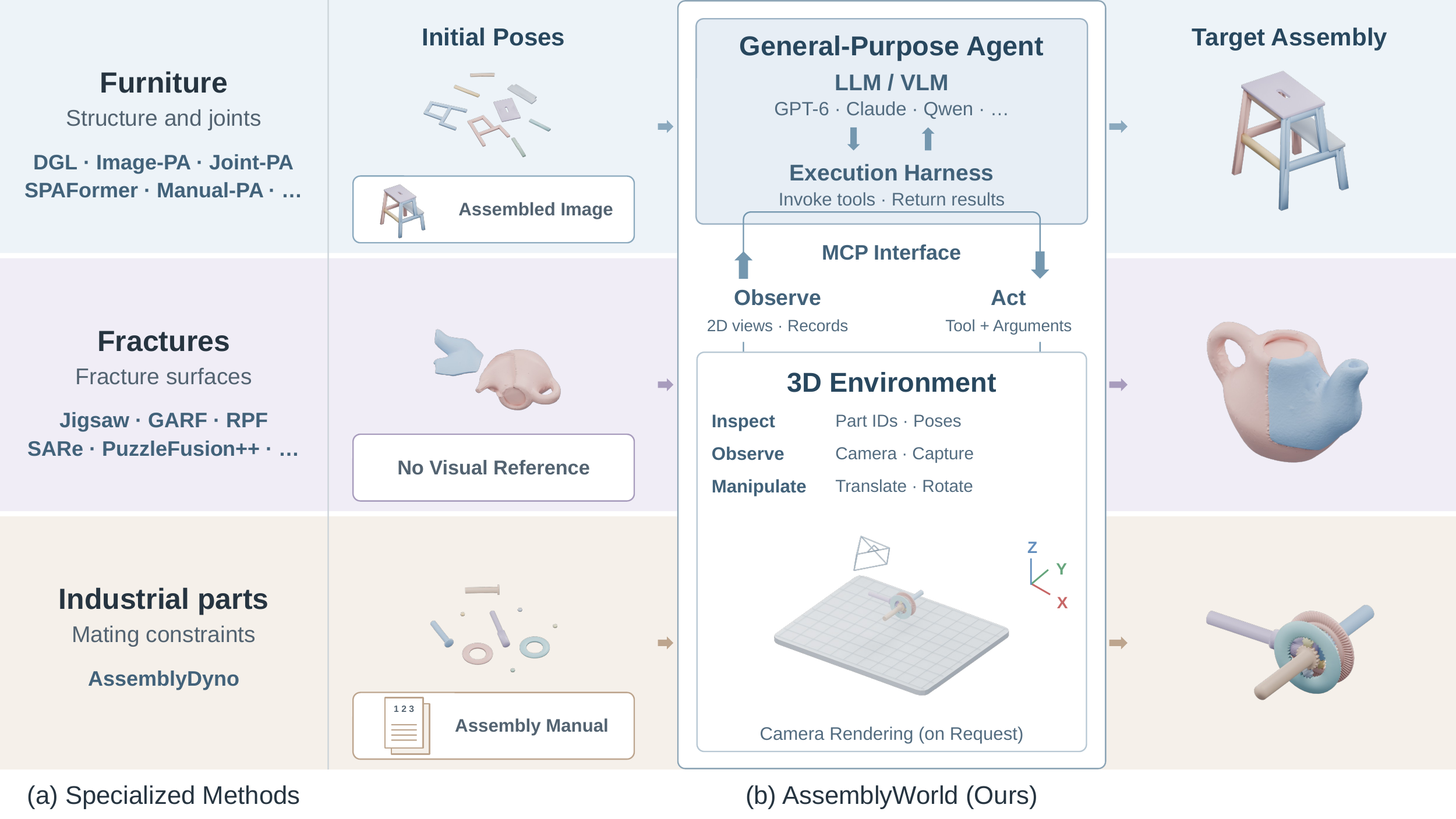}
\captionof{figure}{(a) Prior specialized assembly methods across three domains. (b) AssemblyWorld connects a general-purpose model and execution harness to a 3D environment through Model Context Protocol (MCP). Tools return requested views, scene records, or execution status. References guide the agent; initial parts populate the scene. Targets illustrate assembly goals, not agent outputs. Reference conditions, layouts, colors, and the manual icon are illustrative.}
\label{fig:teaser}
\end{minipage}
\par\vspace{0pt}

\begin{abstract}
The task of 3D assembly requires translating an understanding of parts and their relationships into precise spatial arrangements. Can pretrained general-purpose agents assemble objects through visual interaction without additional assembly-specific fine-tuning? To investigate this question, we introduce \textbf{\AW{}}, an interactive 3D environment in which agents inspect rendered views and manipulate supplied rigid parts, guided by images or assembly manuals when available. Agents perceive part geometry through 2D views rather than direct access to mesh vertices or faces, while their resulting assemblies are evaluated geometrically. Building on this environment, we construct \textbf{\AB{}}, comprising 100 assembly tasks across 80 objects spanning furniture, industrial assembly, and fracture reassembly. Evaluating eight agent systems reveals substantial differences in their capabilities. The strongest system achieves 80.9\% part accuracy but 59.4\% complete-assembly success. The evaluated open-source systems lag substantially behind their stronger closed-source peers in both execution reliability and assembly accuracy. Analyses of visual references, interaction trajectories, and failures show how agents revise assemblies while leaving residual positioning errors. \AW{} provides a common setting for both assessing  the capabilities of interactive assembly agents and characterizing the gap between approximate structure recovery and precise reconstruction.
\end{abstract}

\section{Introduction}
\label{sec:introduction}
Assembling an object requires more than recognizing its parts or understanding what the finished object should look like. A chair may have all its legs beneath the seat and still be incorrectly assembled: connections must align, parts must be correctly oriented, and even small positional errors can leave the object assembly incomplete. These demands arise across domains: furniture assembly~\citep{zhang2025manualpa}, industrial components~\citep{li2026assemblybench}, and fracture reconstruction~\citep{li2025garf}, although the available clues range from well-illustrated manuals to just the geometry of the parts alone.
The task of \emph{3D assembly} therefore provides a concrete test of whether an intelligent system can translate visual and spatial reasoning into precise transformations in a 3D world.

A substantial body of work studies geometric representations~\citep{huang2020dgl,harish2022rgl,du2024hierarchy} and learned pose prediction~\citep{zhang2022iet,cheng2023scorepa,xu2025spaformer,zhao2025assembler,zhang2025cfpa} for 3D part assembly. Reference-guided methods further use assembled-object images or instruction diagrams to inform part placement~\citep{li2020imagepa,wang2025imagine,zhang2025manualpa}.
For example, \citet{zhang2025manualpa} exploit instruction diagrams in Manual-PA, while \citet{li2025garf} model geometric relationships between fragments in GARF. 
General-purpose agents built on models from the OpenAI GPT and Qwen families~\citep{openai2023gpt4,yang2025qwen3} offer a complementary approach. Through tool calls, models specify operations for an execution harness to perform and receive their outputs as context for subsequent decisions. In a 3D environment, this allows an agent to inspect a scene, manipulate parts, and use further observations to revise its decisions. This setting requires choosing informative viewpoints, relating observations across views, and translating an assembly hypothesis into numerical pose edits. Because edits change subsequent observations, the agent must also detect and correct its own errors throughout interaction. Success must therefore be assessed from the assembled geometry, since a plausible verbal explanation need not imply accurate part placement. This raises a central question: \emph{can pretrained general-purpose agents assemble objects through visual interaction without additional assembly-specific fine-tuning?} Answering it requires an environment that connects observations to executable geometric actions and supports comparisons across assembly domains, reference conditions, and agent systems.

Towards this end, we introduce \emph{\AW{}}, an interactive 3D environment for assembling supplied rigid object parts. Within this environment, agents inspect rendered 2D views, choose camera viewpoints, and translate or rotate individual object parts and groups. When available, images of the completed object or assembly manuals provide visual guidance. Since object parts are rigid, agents must solve the task by arranging the given object components. As illustrated in Figure~\ref{fig:teaser}, the same observation-action interface supports furniture, industrial assembly, and fracture reassembly. Geometric evaluation measures accuracy of the resulting configurations, while recorded states make it possible to examine how an assembly develops throughout interaction.

Building on this environment, we construct \emph{\AB{}}, a benchmark comprising 80 objects selected from PartNet~\citep{mo2019partnet}, IKEA-Manual~\citep{wang2022ikea}, AssemblyBench~\citep{li2026assemblybench}, and Fantastic Breaks~\citep{lamb2023fantastic}. Each assembly task consists of an object's parts in a fixed initial configuration and the reference information provided to the agent. We evaluate each of the 20 PartNet objects both with and without an assembled-object image, and each of the remaining 60 objects under its dataset-specific reference setting, yielding 100 assembly tasks across 80 distinct objects. Standardized initial configurations and a common execution budget enable comparisons across eight agent systems. Larger source-level evaluations complement this benchmark by placing agent performance alongside specialized assembly methods. These settings connect cross-domain evaluation with analyses of reference use, interaction dynamics, and residual errors.

Our results show strong assembly capability but a gap between partial progress and complete reconstruction.
The strongest system achieves 80.9\% part accuracy but only 59.4\% complete-assembly success on the common benchmark, while the evaluated open-source systems lag substantially behind their stronger closed-source peers in execution reliability and assembly accuracy. Visual references improve performance for the stronger systems, while trajectories reveal how agents alternate inspection and manipulation, revise intermediate arrangements, and sometimes terminate with unresolved positional errors. Additional experiments examine sensitivity to task variations and the potential for combining interactive agents with specialized geometric models. Together, these findings identify the precision and reliability challenges that remain for general-purpose assembly agents.

\clearpage
\section{Related Work}
\label{sec:related}
\textbf{Learning-Based and Instruction-Guided Assembly.}
Learning-based 3D assembly models relative part poses using geometric and structural cues, including graph- and hierarchy-based reasoning~\citep{huang2020dgl,du2024hierarchy}, generative pose modeling~\citep{cheng2023scorepa}, and visual or instructional guidance~\citep{li2020imagepa,zhang2025manualpa}. Recent language-based methods further incorporate geometric encoding, manual understanding, and constraint reasoning~\citep{jing2026assemlm,tie2025manual2skill,tie2026manual2skillpp}. Fracture reassembly instead relies more heavily on geometric correspondence between fragments~\citep{lu2023jigsaw,li2025garf,sun2025rpf}.  
In contrast, we study whether general-purpose agents can perform these assembly tasks through rendered observations and pose-editing tools without additional assembly-specific fine-tuning.

\textbf{3D Assembly Datasets and Benchmarks.}
Existing datasets capture complementary aspects of the task of 3D assembly. 
PartNet provides hierarchically annotated object parts~\citep{mo2019partnet}, IKEA-Manual pairs furniture geometry with real assembly instructions~\citep{wang2022ikea}, and AssemblyBench extends instruction-guided assembly to industrial objects~\citep{li2026assemblybench}. For fracture reconstruction, Breaking Bad provides simulated fractures~\citep{sellan2022breaking}, while Fantastic Breaks provides scans of physically broken objects~\citep{lamb2023fantastic}. 
\AB{} unifies subsets of these datasets in a common interactive environment with standardized initialization and geometric evaluation, enabling cross-domain agent comparison while retaining the reference conditions used by each dataset.

\textbf{Interactive Agents in 3D Environments.}
Tool-using agents connect language and visual reasoning to actions in executable environments. \citet{liang2023policies} generate action programs in Code as Policies, while \citet{huang2023voxposer} and \citet{huang2024rekep} use spatial objectives and constraints in VoxPoser and ReKep. Related work in 3D content creation includes procedural modeling in 3D-GPT~\citep{sun2025threedgpt} and iterative scene reconstruction through rendered feedback in Thinking in Blender~\citep{he2026blender} and VIGA~\citep{yin2026viga}. In contrast, \AW{} studies interactive assembly of a fixed set of parts through rendered observations and pose-editing tools, with execution-based evaluation following the paradigm of OSWorld~\citep{xie2024osworld}.

\section{\AW} \label{sec:method}
\subsection{Task Formulation}
\label{sec:task}
We study the task of interactive 3D assembly of a fixed set of rigid parts. The task setup provides part meshes $\mathcal{M}=\{M_i\}_{i=1}^{n}$ in a separated initial configuration, together with optional reference information $r$, such as an image of the completed object or an assembly manual. The agent accesses the parts through a 3D environment and uses observations and manipulation actions to determine their final poses $\widehat{T}_i\in\SE$. Part geometry and relative scale remain fixed throughout interaction. The objective is to assemble the supplied parts into a coherent object, following the reference image or instructions when provided.

\subsection{Interactive 3D Environment}
\label{sec:environment}
General-purpose agents interact with external environments through predefined tools. The model selects a tool and generates its arguments; the execution harness invokes the operation and returns its output to the model. In \AW{} (Figure~\ref{fig:teaser}), these tools are exposed through Model Context Protocol (MCP): an agent can request a rendered view, choose a part translation or rotation, and then request another view to inspect the change. The returned information becomes part of the context for subsequent decisions.

\textbf{Observation.}
The agent observes the current assembly through rendered 2D views and can adjust the camera to inspect connections from different viewpoints. Scene-query tools supplement these observations with structured records: \texttt{list\_objects} returns part identifiers, while \texttt{get\_object} reports a selected part's body position, orientation, and mesh-local bounding dimensions. Mesh vertices and faces are hidden; part geometry must be inferred from rendered views.

\textbf{Manipulation and Interaction.}
Tool descriptions specify the available operations and their arguments. To translate or rotate parts, the agent selects their identifiers and specifies a translation vector or rotation angles. It can also choose the coordinate frame and, for rotations, a pivot. Group operations allow several parts to be repositioned together while preserving their relative configuration. The environment executes the requested transformation and returns its status; the agent can then request a rendered view to inspect the result.

The agent determines the sequence of observations and edits and when to finish. We record the operations and resulting states, retaining the final configuration when execution ends.

\subsection{Benchmark Construction}
\label{sec:benchmark}
\textbf{3D Assembly Datasets.}
\AB{} covers three assembly domains using four datasets. PartNet~\citep{mo2019partnet} and IKEA-Manual~\citep{wang2022ikea} provide furniture with semantic parts, such as seats, legs, and tabletops, whose functional roles and spatial relationships help determine the object structure. AssemblyBench~\citep{li2026assemblybench} provides complex industrial assemblies, where components must be arranged according to their geometric fit and mating relationships. Fantastic Breaks~\citep{lamb2023fantastic} contains fractured objects whose fragments provide complementary surface geometry, rather than the semantic part structure (as in furniture assembly).

\textbf{Reference Conditions.}
IKEA-Manual provides real-world IKEA assembly instructions, whereas AssemblyBench provides sequences of rendered assembly diagrams shown from a consistent viewpoint. Fantastic Breaks tasks provide no reference. For PartNet, we evaluate each object both without a reference and with an image of the fully assembled furniture, using the same geometry and initial configuration in both conditions. The paired conditions allow us to compare performance with and without a reference image on the same objects. With only a fixed reference image, the agent must infer occluded part placements from the supplied parts and visible structure. We evaluate the final assembly geometry rather than agreement with an illustrated operation sequence.

\textbf{Standardized Initial Configurations.}
We apply the same initialization procedure across datasets. All parts of an object use a shared scale to preserve their relative dimensions. Each part is placed in a local coordinate frame defined by its principal axes, assigned a randomized yaw, and scattered into a separated initial layout. The normalization scale depends only on individual part geometry. We fix the initial configuration of each object across all evaluated systems.

\textbf{Benchmark Composition.}
We select 20 objects from each dataset through random sampling with part-count stratification where part counts vary. The resulting benchmark provides a common set of assembly tasks for comparing general-purpose agents across domains and reference conditions. The Fantastic Breaks subset contains only two-part objects. Evaluating each PartNet object both with and without a reference image yields 100 assembly tasks across 80 distinct objects. Appendix~\ref{app:runtime} provides task construction and execution details.

\section{Experiments}
\label{sec:experiments}
\subsection{Experimental Setup}
\label{sec:evaluation}
\textbf{Evaluation Settings.}
We compare eight agents on the 100 assembly tasks of AssemblyWorldBench, constructed from the four datasets described in Section~\ref{sec:benchmark}. We also evaluate GPT-6 Astra on larger subsets of these datasets for comparison with specialized methods.

\textbf{Agents and Execution Protocol.}
Table~\ref{tab:benchmark} lists the eight agents and their resource use. GPT, Qwen, and DeepSeek use Codex, with provider adapters for the latter two; Claude models use Claude Code. All systems share the environment tools, initial configurations, task instructions, and 60-minute execution limit. We score the final exported configuration, including partial assemblies after errors or timeouts.

\textbf{Evaluation Metrics.}
For the common benchmark, predicted assemblies are globally aligned to the targets before evaluation, and geometrically equivalent parts are matched using Hungarian assignment.
We report shape Chamfer distance (SCD), part accuracy (PA), and shape success rate (SR). SCD measures the squared bidirectional Chamfer distance between the complete predicted and target assemblies. PA is the fraction of correctly placed parts, where a part is considered correct if its squared bidirectional Chamfer distance to the matched target part is at most $0.01$. SR requires all parts of an object to be correct. 
In our free-space setting, unfinished assemblies may leave parts far from their targets, so a small number of severe failures can dominate the mean SCD.
PA and SR are reported as percentages and SCD as $\times1000$. Overall PA and SR average the four datasets equally, with the two PartNet reference conditions averaged first. The source-level Fantastic Breaks comparison uses a separate anchor-aligned protocol.
Appendices~\ref{app:runtime}--\ref{app:complete_results} provide evaluation details and supplementary quantitative results.

\subsection{Results on \AB{}}
\begin{table}[!t]

\caption{AssemblyWorldBench results: SR (\%) and mean resource use for eight agents. NR, IR, and M denote no reference, an assembled-object image, and a manual. IKEA, AB, and FB denote IKEA-Manual, AssemblyBench, and Fantastic Breaks. Bold marks the highest SR or lowest resource use.}
\label{tab:benchmark}
\centering
\footnotesize
\setlength{\tabcolsep}{2pt}
\begin{NiceTabular*}{\linewidth}{@{\extracolsep{\fill}}ll!{\hspace{2pt}}*{6}{c}|*{4}{c}@{}}
\CodeBefore
\rowcolor{gray!12}{10}
\Body
\toprule
\Block{2-1}{LLM} & \Block{2-1}{Harness} & \Block{1-2}{PartNet} & & IKEA & AB & FB & Overall & Time & MCP & Tokens & Cost \\
\cmidrule(lr){3-4}\cmidrule(lr){5-5}\cmidrule(lr){6-6}\cmidrule(lr){7-7}
& & NR & IR & M & M & NR & SR (\%) & (min) & calls & (M) & (\$) \\
\midrule
DeepSeek V4.1 Flash & Codex & 0.000 & 0.000 & 0.000 & 0.000 & 0.000 & 0.000 & 26.7 & 186 & 2.59 & \textbf{0.16} \\
Qwen3.8 Max & Codex & 5.000 & 0.000 & 0.000 & 10.00 & 35.00 & 11.90 & 54.5 & 107 & 3.67 & 1.67 \\
\midrule
Claude Sonnet 5 & Claude Code & 0.000 & 0.000 & 0.000 & 5.000 & 25.00 & 7.500 & 20.6 & 196 & 16.45 & 4.72 \\
Claude Opus 5 & Claude Code & 10.00 & 25.00 & 50.00 & 40.00 & 70.00 & 44.40 & 37.0 & 180 & 16.89 & 12.38 \\
Claude Fable 5.1 & Claude Code & 5.000 & 25.00 & 50.00 & 45.00 & \textbf{90.00} & 50.00 & 22.1 & 162 & 4.60 & 6.84 \\
\midrule
GPT-5.6 Terra & Codex & 0.000 & 0.000 & 0.000 & 0.000 & 30.00 & 7.500 & \textbf{4.0} & \textbf{60} & 1.32 & 0.50 \\
GPT-5.6 Sol & Codex & 0.000 & 0.000 & 5.000 & 10.00 & 30.00 & 11.20 & 8.7 & 208 & 2.77 & 1.79 \\
GPT-6 Astra & Codex & \textbf{15.00} & \textbf{40.00} & \textbf{65.00} & \textbf{55.00} & \textbf{90.00} & \textbf{59.40} & 5.8 & 94 & \textbf{1.08} & 2.02 \\
\bottomrule
\end{NiceTabular*}
\end{table}

\paragraph{Assembly Quality.}
Table~\ref{tab:benchmark} compares assembly quality and resource use across the eight systems. Astra achieves the highest overall SR at 59.4\%, followed by Fable at 50.0\% and Opus at 44.4\%. Performance nevertheless varies across domains: Astra and Fable both reach 90\% SR on two-part fracture reassembly, but only 40\% and 25\%, respectively, on image-conditioned PartNet tasks. High success on fragment alignment therefore does not extend uniformly to multi-part furniture assembly.

Qwen reaches 35\% SR on two-part fracture reassembly, but its overall SR is 11.9\%; DeepSeek completes no task successfully. Of their 100 evaluations, 65 and 17 reach the time limit, while another 14 and 34 end in execution failures. These results reflect the deployed model--harness combinations, including execution reliability as well as assembly accuracy.

\textbf{Time and Computational Cost.}
Higher success does not necessarily require more time or tool calls. Astra averages 5.8 minutes and 94 calls per evaluation, versus Sol's 8.7 minutes and 208 calls, while achieving substantially higher SR. Astra also uses less time and recorded cost than Fable and Opus, whereas Terra uses fewer resources but reaches only 7.5\% SR.

\subsection{Comparison with Specialized Assembly Methods}
Tables~\ref{tab:partnet}--\ref{tab:fantastic} compare Astra with specialized assembly methods on larger source-level evaluation sets, using the reported settings of each method.

\begin{table}[!t]

\caption{Assembly results on PartNet. Astra uses no target reference (--) or an image of the assembled furniture; scene observation is available in both conditions.}
\label{tab:partnet}
\centering
\footnotesize
\setlength{\tabcolsep}{2pt}
\begin{NiceTabular*}{\linewidth}{@{\extracolsep{\fill}}ll*{9}{r}@{}}
\CodeBefore
\rowcolor{gray!12}{13,18}
\Body
\toprule
\Block{2-1}{Method} & \Block{2-1}{Condition} & \Block{1-3}{SCD $\downarrow$} & & & \Block{1-3}{PA $\uparrow$} & & & \Block{1-3}{SR $\uparrow$} & & \\
\cmidrule(lr){3-5}\cmidrule(lr){6-8}\cmidrule(lr){9-11}
 & & Chair & Table & Storage & Chair & Table & Storage & Chair & Table & Storage \\
\midrule
\venue{DGL}{NeurIPS'20} & -- & 9.1 & 5.0 & -- & 39.00 & 49.51 & -- & -- & -- & -- \\
\venue{RGL}{WACV'22} & Sequence & 8.7 & 4.8 & -- & 49.06 & 54.16 & -- & -- & -- & -- \\
\venue{IET}{RA-L'22} & -- & 5.4 & 3.5 & -- & 62.80 & 61.67 & -- & -- & -- & -- \\
\venue{Score-PA}{BMVC'23} & -- & 7.4 & 4.5 & -- & 42.11 & 51.55 & -- & 8.320 & 11.23 & -- \\
\venue{CCS}{AAAI'24} & -- & 7.0 & -- & -- & 53.59 & -- & -- & -- & -- & -- \\
\venue{3DHPA}{CVPR'24} & -- & 5.1 & \textbf{2.8} & -- & 64.13 & 64.83 & -- & -- & -- & -- \\
\venue{Joint-PA}{CVPR'24} & Joint & 6.0 & 7.0 & -- & \textbf{72.80} & 67.40 & -- & -- & -- & -- \\
\venue{SPAFormer}{3DV'25} & Sequence & 6.7 & 3.8 & \textbf{4.5} & 55.88 & 64.38 & \textbf{56.11} & 16.40 & 33.50 & 7.850 \\
\venue{CFPA}{NeurIPS'25} & -- & \textbf{4.9} & 3.3 & -- & 69.24 & 68.48 & -- & -- & -- & -- \\
\venue{Assembler}{SA'25} & -- & 9.2 & 8.0 & -- & 61.85 & 65.04 & -- & 22.59 & 39.07 & -- \\
GPT-6 Astra + Codex & -- & 69.9 & 47.0 & 141.6 & 55.68 & \textbf{71.39} & 50.06 & \textbf{22.63} & \textbf{48.97} & \textbf{12.84} \\
\midrule
\venue{Image-PA}{ECCV'20} & Image & 6.7 & 3.7 & 5.0 & 45.40 & 71.60 & 40.20 & -- & -- & -- \\
\venue{Manual-PA}{ICCV'25} & Image & 5.9 & 3.9 & 3.7 & 62.67 & 70.10 & 47.79 & 19.97 & 32.83 & 4.730 \\
\venue{Imagine}{AAAI'25} & Image & \textbf{4.2} & \textbf{2.5} & \textbf{2.4} & 65.88 & 65.13 & 56.35 & -- & -- & -- \\
\venue{Assembler}{SA'25} & Image & 7.2 & 4.7 & -- & 66.17 & 70.44 & -- & 24.92 & 43.46 & -- \\
GPT-6 Astra + Codex & Image & 29.5 & 21.0 & 184.8 & \textbf{76.45} & \textbf{85.35} & \textbf{63.87} & \textbf{43.36} & \textbf{63.60} & \textbf{20.27} \\
\bottomrule
\end{NiceTabular*}
\vspace{7pt}
\end{table}

\textbf{Furniture Assembly.}
PartNet methods commonly train category-specific models, including Manual-PA, Imagine~\citep{wang2025imagine}, and Assembler~\citep{zhao2025assembler}. Cross-category transfer is studied by Manual-PA and Imagine, and multi-category training by SPAFormer~\citep{xu2025spaformer} and Assembler. We use the same pretrained Astra agent across furniture categories without additional assembly-specific fine-tuning or category-specific models.

With image guidance, Astra exceeds the compared methods in PA and SR across all three PartNet categories (Table~\ref{tab:partnet}), reaching 85.35\% PA and 63.60\% SR on tables. Removing the reference reduces PA in every category. Image-conditioned Storage reaches 63.87\% PA but only 20.27\% SR, illustrating the gap between correctly placing individual parts and completing an assembly. Free-space failures can nevertheless dominate SCD: 8 of 148 image-conditioned Storage objects (5.4\%) contribute 99.3\% of summed SCD, with six retaining all initial part poses. Their separated parts yield large squared distances, unlike predictors with bounded translations such as DGL~\citep{huang2020dgl}.

\begin{table}[!t]

\caption{Assembly results on IKEA-Manual. Subset columns use equal-category averages. PA and SR are percentages; SR requires all parts correct, whereas $\mathrm{SR}^{a}$ uses SCD $<0.02$ before scaling. $\dagger$ denotes Image-PA retrained on diagrams.}
\label{tab:ikea}
\centering
\footnotesize
\setlength{\tabcolsep}{2pt}
\begin{NiceTabular*}{\linewidth}{@{\extracolsep{\fill}}l*{7}{r}@{}}
\CodeBefore
\rowcolor{gray!12}{8}
\Body
\toprule
\Block{2-1}{Method} & \Block[l]{1-3}{Chair/Table} & & & \Block[l]{1-2}{Bench/Chair/Desk} & & \Block[l]{1-2}{Full} & \\
\cmidrule(lr){2-4}\cmidrule(lr){5-6}\cmidrule(lr){7-8}
 & SCD $\downarrow$ & PA $\uparrow$ & SR $\uparrow$ & SCD $\downarrow$ & $\mathrm{SR}^{a}$ $\uparrow$ & SCD $\downarrow$ & $\mathrm{SR}^{a}$ $\uparrow$ \\
\midrule
\venue{3DHPA}{CVPR'24} & 36.1 & 2.971 & 0.000 & -- & -- & -- & -- \\
\venue{Image-PA}{ECCV\textquotesingle20}$^\dagger$ & 16.0 & 27.91 & 5.265 & -- & -- & -- & -- \\
\venue{Manual-PA}{ICCV'25} & 8.1 & 46.12 & 9.650 & -- & -- & 21.6 & 54.90 \\
\venue{TwoByTwo}{CVPR'25} & -- & -- & -- & 221.7 & 6.500 & -- & -- \\
\venue{AssemLM}{arXiv'26} & -- & -- & -- & 21.3 & 81.00 & 61.3 & 69.60 \\
GPT-6 Astra + Codex & \textbf{0.8} & \textbf{93.27} & \textbf{75.44} & \textbf{1.0} & \textbf{100.0} & \textbf{0.9} & \textbf{100.0} \\
\bottomrule
\end{NiceTabular*}
\vspace{7pt}
\end{table}

\begin{figure}[!t]
\centering
\begin{minipage}[t]{0.498\linewidth}
\centering\scriptsize
\textcolor[HTML]{99a6b0}{\rule[.5ex]{9pt}{1pt}} Inspect\quad
\textcolor[HTML]{397ca5}{\rule[.5ex]{9pt}{1pt}} Observe\quad
\textcolor[HTML]{df9957}{\rule[.5ex]{9pt}{1pt}} Manipulate
\end{minipage}\hfill\begin{minipage}[t]{0.498\linewidth}
\centering\scriptsize
\textcolor[HTML]{1769aa}{\rule[.5ex]{9pt}{1pt}} Astra\quad
\textcolor[HTML]{de7722}{\rule[.5ex]{9pt}{1pt}} Fable
\end{minipage}\par
\vspace{0pt}
\begin{minipage}[t]{0.247\linewidth}
\centering
\includegraphics[width=\linewidth]{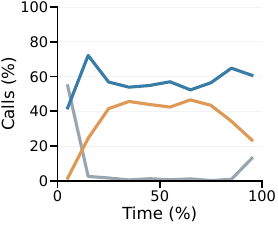}\\[-2pt]
{\footnotesize (a) Astra activity}
\end{minipage}\hfill\begin{minipage}[t]{0.247\linewidth}
\centering
\includegraphics[width=\linewidth]{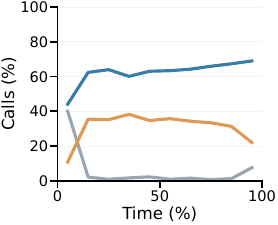}\\[-2pt]
{\footnotesize (b) Fable activity}
\end{minipage}\hfill\begin{minipage}[t]{0.247\linewidth}
\centering
\includegraphics[width=\linewidth]{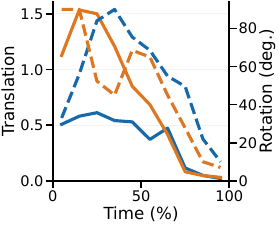}\\[-2pt]
{\footnotesize (c) Manipulation}
\end{minipage}\hfill\begin{minipage}[t]{0.247\linewidth}
\centering
\includegraphics[width=\linewidth]{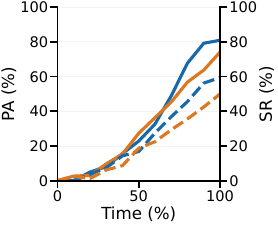}\\[-2pt]
{\footnotesize (d) Assembly quality}
\end{minipage}
\caption{Interaction dynamics of Astra and Fable across 100 evaluations each. (a--b) Mean call shares within active evaluation-bins. (c) Median per-evaluation translation and rotation magnitudes; translation is normalized by the largest-part diagonal. (d) Mean PA and SR. All four plots use benchmark source weights and normalized interaction time. In (c--d), colors identify systems; solid and dashed lines use the left and right axes, respectively.}
\label{fig:behavior}
\vspace{6pt}
\end{figure}

Real manuals add challenges beyond rendered diagrams: Manual-PA uses synthetic stepwise diagrams for PartNet and IKEA-Manual and identifies arrows, close-ups, part labels, and hierarchical subassemblies as obstacles to transfer. Manual2Skill handles real manuals with graph extraction and learned pose estimation. Astra instead uses the original IKEA pages to assemble through interaction, without additional assembly-specific fine-tuning or a dedicated pose predictor.

On IKEA-Manual (Table~\ref{tab:ikea}), Astra achieves category-averaged PA/SR of 93.27\%/75.44\% on Chair/Table, exceeding the compared methods. It also obtains lower SCD and higher $\mathrm{SR}^{a}$ on Full and Bench/Chair/Desk, with full-dataset SCD of 0.9 and 100\% $\mathrm{SR}^{a}$ in both settings. These results show that a pretrained general-purpose agent can interpret real instruction manuals and construct furniture geometry through interaction, without an assembly-specific pose predictor.

\Needspace{13\baselineskip}
\begin{wraptable}{r}{0.52\textwidth}

\caption{
Assembly results on AssemblyBench. $N$ counts evaluated objects; PA/SR are percentages and SCD is scaled by 1000. Astra is evaluated on 279 of 280 objects, excluding one refusal involving a dangerous item.
}
\label{tab:assemblybench}
\centering
\footnotesize
\setlength{\tabcolsep}{3pt}
\begin{NiceTabular}{@{}lrrrr@{}}
\CodeBefore
\rowcolor{gray!12}{4}
\Body
\toprule
Method & $N$ & SCD $\downarrow$ & PA $\uparrow$ & SR $\uparrow$ \\
\midrule
\venue{Manual-PA}{ICCV'25} & 280 & 4.24 & 70.04 & 33.57 \\
\venue{AssemblyDyno}{CVPR'26} & 280 & \textbf{3.91} & 71.21 & 34.64 \\
\midrule
Astra & 279 & 8.54 & \textbf{78.32} & \textbf{55.91} \\
\bottomrule
\end{NiceTabular}
\end{wraptable}

\textbf{Industrial Assembly.}
Industrial assembly introduces substantial differences in relative part size within a single object. The median largest-to-smallest part-size ratio is 4.39 on AssemblyBench, compared with 2.20 on IKEA-Manual and 1.79 on PartNet. The ratio of per-part PCA (principal component analysis) bounding-box diagonals is scale-invariant. Using stepwise rendered instructions, the same pretrained Astra agent achieves 78.32\% PA and 55.91\% SR across 279 AssemblyBench objects without additional assembly-specific fine-tuning, exceeding the reported results of the specialized methods in Table~\ref{tab:assemblybench}.

\vspace{8pt}
\Needspace{18\baselineskip}
\begin{wraptable}{r}{0.52\textwidth}

\caption{Assembly results on Fantastic Breaks. RE/TE are rotation/translation errors. Units: RE in degrees, PA in percent, TE/CD scaled by 100/1000. RE is Euler-angle root mean square error (RMSE) for Jigsaw, PF++, GARF, and Astra; geodesic otherwise. Astra uses anchor alignment on 150 objects.}
\label{tab:fantastic}
\centering
\footnotesize
\setlength{\tabcolsep}{3pt}
\begin{NiceTabular}{@{}lrrrr@{}}
\CodeBefore
\rowcolor{gray!12}{8}
\Body
\toprule
Method & RE $\downarrow$ & TE $\downarrow$ & PA $\uparrow$ & CD $\downarrow$ \\
\midrule
\venue{Jigsaw}{NeurIPS'23} & $26.30^\circ$ & 6.43 & 73.64 & 10.47 \\
\venue{PF++}{ICLR'25} & $20.68^\circ$ & 4.37 & 83.33 & 6.68 \\
\venue{GARF}{ICCV'25} & $10.62^\circ$ & 2.10 & 91.00 & 2.12 \\
\venue{RPF}{NeurIPS'25} & $6.32^\circ$ & 2.18 & 96.90 & 2.53 \\
\venue{TORA-CKA}{ECCV'26} & $\mathbf{3.03}^\circ$ & 0.80 & 97.28 & 0.26 \\
\venue{SARe-Gen}{arXiv'26} & $7.78^\circ$ & \textbf{0.35} & \textbf{98.53} & \textbf{0.23} \\
\midrule
Astra & $11.45^\circ$ & 2.54 & 91.67 & 6.98 \\
\bottomrule
\end{NiceTabular}
\end{wraptable}

\textbf{Fracture Reassembly.}
In our fracture reassembly setting, agents receive no visual reference of the target object, such as an image of its intact shape. Within \AW's interactive 3D environment, the agent cannot directly access mesh vertices, faces, or point clouds; it must infer how fragments fit together and adjust their poses using feedback from rendered 2D views. Despite this observation constraint, Astra achieves 91.67\% PA on 150 Fantastic Breaks objects without additional fracture-specific fine-tuning. Table~\ref{tab:fantastic} compares its performance with specialized methods using published results reported by \citet{li2025garf} for GARF and \citet{jia2026sare} for SARe. The higher PA and lower translation errors reported by recent specialized methods highlight their strength in precise fragment placement.

\vspace{8pt}
\subsection{Analysis of Interactive Assembly}
\label{sec:performance_analysis}
\begin{figure}[!t]
\centering
\includegraphics[width=\linewidth]{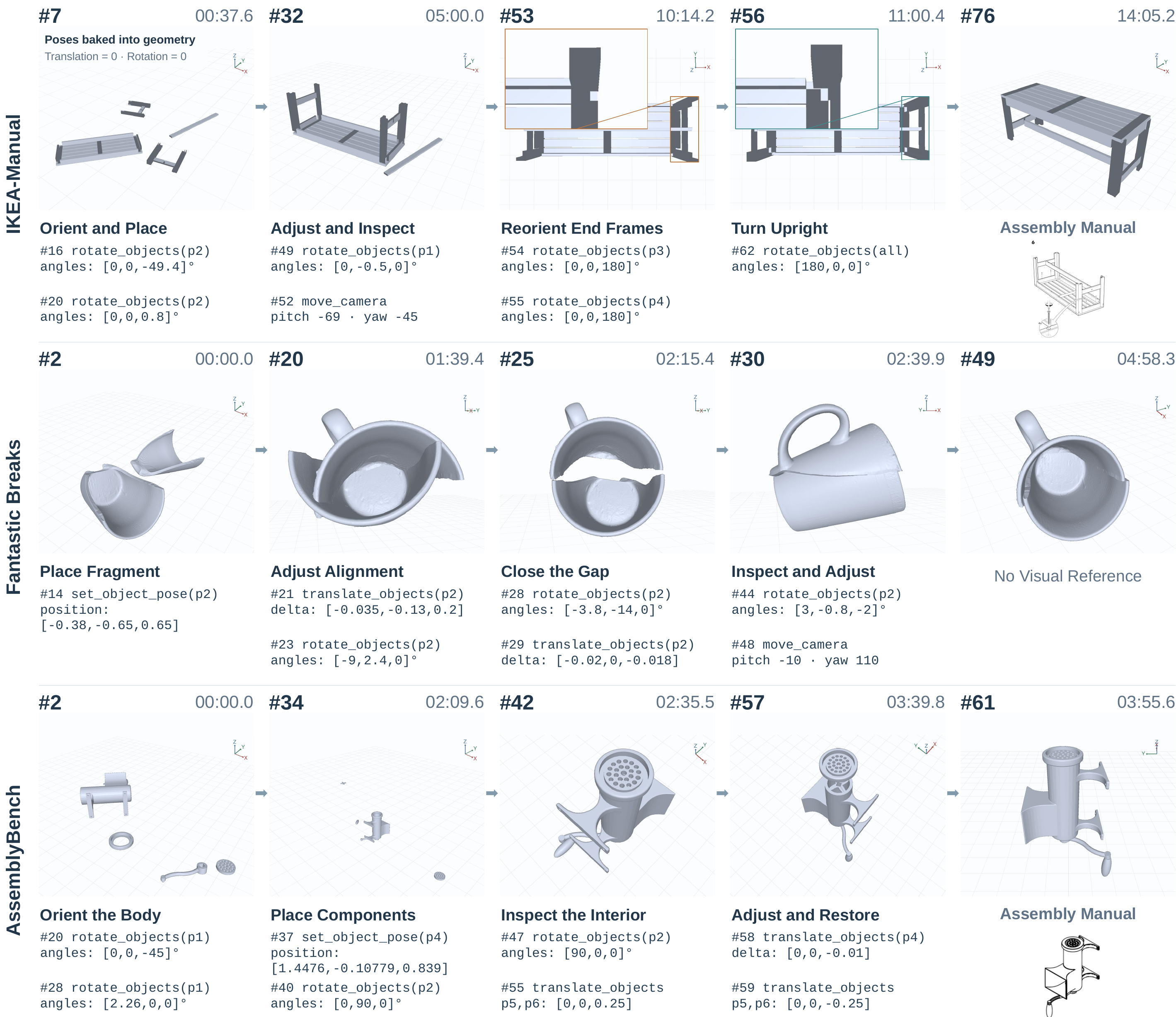}
\caption{
GPT-6 Astra interaction trajectories across three assembly domains. Recorded states are re-rendered from the corresponding camera views. Labels show environment-call numbers (\#) and elapsed time; selected MCP calls connect successive views. Insets highlight the IKEA end-frame correction. Part IDs and arguments are abbreviated; angles are in degrees.
}
\label{fig:interaction}
\end{figure}

\textbf{Interaction Dynamics.}
Agents actively inspect intermediate assemblies and adjust part poses using visual feedback. Figure~\ref{fig:interaction} shows Astra changing viewpoint to examine the bench end frames and correcting their orientation before turning the assembly upright. In the industrial example, it temporarily lifts the perforated plate and retaining ring to inspect and adjust the blade beneath them, then restores the lifted parts. Such inspection can require moving parts away from their target poses, so intermediate geometric accuracy alone does not determine whether an action is useful.

Figure~\ref{fig:behavior} summarizes how these interactions evolve across Astra and Fable trajectories. In Figure~\ref{fig:behavior}(a--b), inspection queries are concentrated at the beginning, when agents acquire scene and part information. Manipulation then becomes more frequent alongside sustained visual observation. Toward the end, manipulation decreases while observation remains prominent, consistent with a shift toward checking the resulting configuration. Figure~\ref{fig:behavior}(c) shows an accompanying reduction in translation and rotation magnitudes, while Figure~\ref{fig:behavior}(d) shows overall increases in mean PA and SR. Manipulation magnitudes also decrease in failed evaluations, so smaller edits alone do not establish convergence to a correct assembly.
\begin{wrapfigure}{R}{0.36\textwidth}
\centering
\includegraphics[width=\linewidth]{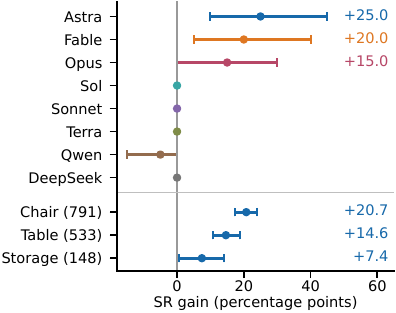}
\caption{SR gains from reference images, with paired 95\% intervals: benchmark (top) and Astra on larger PartNet sets (bottom). Labels give gains in percentage points.}
\label{fig:reference_analysis}
\end{wrapfigure}

\paragraph{Effect of Visual References.}
Reference images improve assembly accuracy, but their benefit varies across systems. Figure~\ref{fig:reference_analysis} compares paired evaluations with and without an image of the assembled furniture. On the benchmark's 20 PartNet objects, Astra, Fable, and Opus gain 25, 20, and 15 SR points, respectively; the other systems show no gain or a decrease. The larger PartNet evaluations in Table~\ref{tab:partnet} show a similar benefit for Astra. Across 533 tables and 148 storage objects, SR increases by 14.6 and 7.4 points, respectively. These gains extend beyond the benchmark subset and differ between furniture categories.

\begin{figure}[!t]
\centering
\begin{minipage}[t]{0.325\linewidth}
\centering
\includegraphics[width=\linewidth]{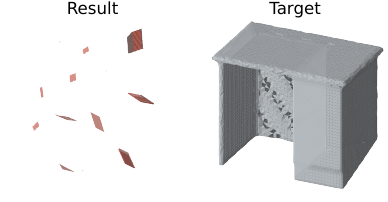}\\[-2pt]
{\footnotesize (a) Terra}
\end{minipage}\hfill\begin{minipage}[t]{0.325\linewidth}
\centering
\includegraphics[width=\linewidth]{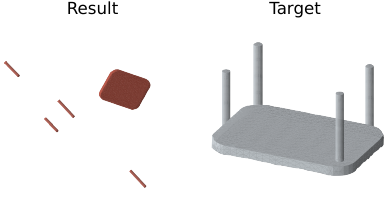}\\[-2pt]
{\footnotesize (b) Qwen}
\end{minipage}\hfill\begin{minipage}[t]{0.325\linewidth}
\centering
\includegraphics[width=\linewidth]{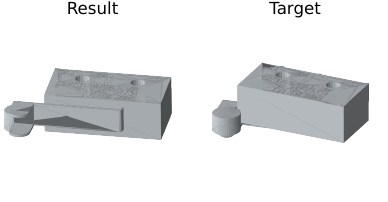}\\[-2pt]
{\footnotesize (c) Astra}
\end{minipage}
\caption{Recorded failure cases and targets. (a) Logical grouping leaves parts dispersed. (b) Numerical pose placement leaves components detached. (c) Eight of nine parts are correct, with one residual positional error. Red marks incorrect parts; teal marks the matched target in (c). Views in (a--b) are fitted independently.}
\label{fig:failure_completion}
\end{figure}

\Needspace{6\baselineskip}
\textbf{Failure Modes.}
Failures span execution and geometry: Qwen and DeepSeek records contain unsupported-tool and stream-disconnection errors, while all 49 normally terminated DeepSeek runs fail geometrically. Figure~\ref{fig:failure_completion}(a--b) shows Terra grouping 13 dispersed parts and Qwen reporting numerical alignment despite detached components. These cases reveal difficulties in assessing visible geometry from pose readback.

Some Astra and Fable failures retain much of the target structure: translation-dominated errors account for 67.4\% and 75.1\% of their incorrect parts. These parts pass the correctness threshold after centroid alignment without changing orientation. Figure~\ref{fig:failure_completion}(c) shows an Astra industrial assembly with eight of nine parts correct and one positional error; a Fable case has six of seven correct. Their reports describe unresolved seating or internal fits inferred from rendered views.

These residual errors motivate combining visual interaction with geometric refinement, while checking whether either composition degrades correct structure. Appendix~\ref{app:behavior} extends the interaction and failure analyses.

\begin{wraptable}{R}{0.48\textwidth}
\centering
\caption{Bidirectional refinement on 150 Fantastic Breaks objects with shared point samples and anchor-aligned evaluation. Arrows indicate execution order. RE is in degrees, PA in percent, and TE/CD scaled by 100/1000.}
\label{tab:hybrid}
\footnotesize
\setlength{\tabcolsep}{3pt}
\begin{NiceTabular}{@{}lrrrr@{}}
\toprule
Method & RE $\downarrow$ & TE $\downarrow$ & PA $\uparrow$ & CD $\downarrow$ \\
\midrule
Agent & 11.45 & 2.54 & 91.67 & 6.98 \\
GARF & 14.78 & 5.16 & 83.67 & 8.82 \\
Agent $\rightarrow$ GARF & \textbf{9.34} & 4.24 & 87.00 & 6.94 \\
GARF $\rightarrow$ Agent & 10.35 & \textbf{1.84} & \textbf{93.00} & \textbf{4.35} \\
\bottomrule
\end{NiceTabular}
\end{wraptable}

\textbf{Combining Agents with Geometric Models.}
\label{sec:hybrid_refinement}
Table~\ref{tab:hybrid} compares both composition orders of Astra and GARF on 150 two-fragment Fantastic Breaks objects with shared point samples and metrics. Agent initialization raises GARF PA from 83.67\% to 87.00\%, below the agent-only 91.67\%: refinement reduces rotation error but increases translation error, degrading some correct assemblies. GARF was not trained to refine agent outputs and is applied without retraining.

Conversely, Astra inspects GARF predictions through rendered views and adjusts their part poses, reaching 93.00\% PA and reducing CD from $8.82\times10^{-3}$ to $4.35\times10^{-3}$. All four mean metrics improve over the agent-only baseline. These results support visual correction of geometric predictions and motivate refinement adapted to agent-generated initializations. Appendix~\ref{app:refinement} reports the matched protocol and paired changes.

\textbf{Robustness to Task Variations.}
\label{sec:robustness}
We evaluate Astra's sensitivity to initial layouts and part-set changes on five fixed objects per benchmark block, with paired PartNet reference conditions and one run per condition. Aggregate SR changes little across layouts, from 52.5\% to 52.5\% and 50.0\%, but some objects switch between success and failure. Similar aggregate performance does not imply consistent per-object success.

With one part removed, retained-part PA differs substantially across domains: 88.4\% on IKEA-Manual versus 43.3\% on AssemblyBench. With an added distractor, Astra sometimes leaves it unincorporated and sometimes attempts to assign it a role. Because the unchanged prompt requests that all supplied parts be incorporated, these behaviors cannot be interpreted directly as anomaly-detection ability. Appendix~\ref{app:robustness} provides the complete paired results and trajectory audits. These small-sample, single-run findings remain exploratory.

\FloatBarrier
\section{Conclusion}
\label{sec:conclusion}
\label{sec:limitations}
We introduced \AW{}, an interactive 3D assembly environment, and \AB{}, a cross-domain benchmark evaluating general-purpose agents with a common interface and geometric criteria. Across eight systems, Astra achieves the highest overall success rate, while the stronger closed-source systems substantially outperform the evaluated open-source systems. The same pretrained agent can assemble furniture, industrial objects, and fractured objects without additional assembly-specific fine-tuning. The gap between part accuracy and complete-assembly success nevertheless shows that broad capability does not ensure geometric precision.

These findings motivate a division of labor in which general-purpose models provide cross-domain reasoning, while specialized geometric methods provide precise pose estimation and refinement. The dependence on composition order in our refinement study highlights the need to design this interaction jointly. Our evaluation concerns free-space geometry and does not establish collision-free or physically stable assembly. The sampled tasks, single-run robustness tests, and two-fragment, single-seed refinement study limit the scope of our findings; unknown pretraining exposure also prevents claims of contamination-free generalization. Within these bounds, \AW{} provides a common platform for investigating how general-purpose reasoning and specialized geometry can jointly improve assembly across domains.

\clearpage
\bibliography{references}
\bibliographystyle{iclr}
\clearpage
\appendix
\setcounter{figure}{0}
\setcounter{table}{0}
\renewcommand{\thefigure}{A-\arabic{figure}}
\renewcommand{\thetable}{A-\arabic{table}}
\renewcommand{\theHfigure}{appendix.\arabic{figure}}
\renewcommand{\theHtable}{appendix.\arabic{table}}
\section{Experimental Details}
\label{app:runtime}
\label{app:metrics}
\subsection{Datasets and Task Construction}
\label{app:selection}
Table~\ref{tab:dataset_coverage} distinguishes the common benchmark from the larger source-level evaluations. The benchmark contains 20 objects per source. PartNet objects have paired image and no-reference conditions, giving 100 tasks across 80 distinct objects. All systems receive the same initial configurations and references.
\begin{table}[!htbp]
\centering\footnotesize
\caption{Dataset coverage in the reported evaluations. Upper rows are the subsets used for all eight systems; lower rows are the larger source-level evaluations. Counts describe evaluation sets, not full datasets; parts are counted once per object, irrespective of reference condition or system. The paired PartNet benchmark conditions give 40 assembly tasks from 20 objects. AssemblyBench retains 279 of 280 objects.}
\label{tab:dataset_coverage}
\begin{NiceTabular*}{\linewidth}{@{\extracolsep{\fill}}lrrrl@{}}
\toprule
Dataset / evaluation set & Objects & Total parts & Parts/object & Reference \\
\midrule
\multicolumn{5}{l}{\textit{Benchmark subsets}} \\
PartNet & 20 & 218 & 5--15 & None / assembled-furniture image \\
IKEA-Manual & 20 & 163 & 3--19 & Real IKEA instructions \\
AssemblyBench & 20 & 139 & 3--18 & Rendered step diagrams \\
Fantastic Breaks & 20 & 40 & 2 & None \\
\midrule
\multicolumn{5}{l}{\textit{Larger evaluations}} \\
PartNet chair & 791 & 8,596 & 2--20 & None / assembled-furniture image \\
PartNet table & 533 & 4,732 & 2--20 & None / assembled-furniture image \\
PartNet storage & 148 & 1,970 & 5--20 & None / assembled-furniture image \\
IKEA-Manual & 102 & 754 & 2--19 & Real IKEA instructions \\
AssemblyBench & 279 & 1,821 & 3--20 & Rendered step diagrams \\
Fantastic Breaks & 150 & 300 & 2 & None \\
\bottomrule
\end{NiceTabular*}
\end{table}

\paragraph{Sampling.}
Objects are randomly sampled with part-count stratification and category coverage. Fantastic Breaks contains only two-part objects and is sampled by category without part-count stratification. These subsets support common-task comparisons across systems rather than estimates of full-dataset performance.

One AssemblyBench object was replaced after content screening, preserving its original category and part-count band. Two additional AssemblyBench objects were excluded from the candidate pool during screening. The larger AssemblyBench evaluation retains 279 of 280 objects after a dangerous-item refusal and uses a successful retry for one other object. Its reported means are conditional on these choices. The original attempt for that retry is unavailable. These source-level results are distinct from the 100-task benchmark aggregate.

The larger PartNet comparisons use the same objects with and without an assembled-object image. IKEA-Manual comprises 57 chairs, 19 tables, eight benches, four desks, three shelves, and 11 miscellaneous objects; Chair/Table and Bench/Chair/Desk summaries average category means equally.

\paragraph{Initialization.}
\label{app:preparation}
Each part is expressed in a right-handed vertex-PCA frame, with its smallest principal axis along local $Z$. A shared scale, twice the largest vertex radius about any part's centroid, preserves relative part sizes without using assembled poses. Parts receive independent yaw angles sampled uniformly from $[-\pi,\pi)$ and are scattered on the XY plane with non-overlapping bounding boxes and a minimum gap of $0.02$ normalized units. Sampling and initialization use seed zero. Initial placements are baked into the mesh vertices, so all initial body transforms are identity transforms. Body origins therefore need not coincide with visual part centers.

Part identifiers are anonymized. Target poses, evaluator point clouds, and equivalence annotations are unavailable to the agent. Reference images are supplied separately: none for no-reference tasks, the final illustrated manual page for PartNet image conditions, all published pages for IKEA-Manual, and one rendered diagram per step for AssemblyBench.

\subsection{Agents, Tools, and Instructions}
All systems access the same MCP tools (Table~\ref{tab:tools}); the harness executes model-selected operations and returns their outputs. Pose edits return execution status, requiring a separate observation to inspect their effect.
\begin{table}[t]
\caption{Environment tools shared by all evaluated systems. Coordinates use normalized scene units with world up $+Z$; quaternions use $(w,x,y,z)$. Group identifiers select their constituent parts.}
\label{tab:tools}
\centering\scriptsize
\setlength{\tabcolsep}{3pt}
\begin{NiceTabular}{@{}l>{\raggedright\arraybackslash}p{0.65\linewidth}@{}}
\toprule
Tools & Information or operation \\
\midrule
\texttt{list\_objects} & Lists object and group identifiers. \\
\texttt{get\_object}, \texttt{get\_scene}, \texttt{get\_state} & Reports body poses, local bounding extents, scene conventions, groups, and camera state. Mesh vertices and faces are unavailable. \\
\texttt{capture\_scene} & Returns a $1024\times768$ image with a $38^\circ$ vertical field of view. \\
\texttt{move\_camera} & Orbits, zooms, or pans using yaw, pitch, zoom, and lateral/vertical offsets. \\
\texttt{translate\_objects} & Applies a translation vector in world or camera coordinates. \\
\texttt{rotate\_objects} & Applies $X$, $Y$, then $Z$ angles in degrees, in world or camera coordinates, about a specified pivot. The default pivot is the mean of selected body origins. \\
\texttt{set\_object\_pose} & Sets a body's position and/or quaternion directly. \\
\texttt{group\_objects}, \texttt{ungroup\_objects} & Creates or dissolves transformation groups without changing relative part poses or snapping parts together. \\
\texttt{start\_episode} & Marks the interaction as active. \\
\bottomrule
\end{NiceTabular}
\end{table}

Astra, Sol, and Terra use Codex CLI 0.155.1 with medium reasoning effort. Fable, Opus, and Sonnet use Claude Code 2.1.272 with medium effort. Qwen uses Codex 0.154.0 and DeepSeek uses Codex 0.155.0, both with the recorded xhigh setting and provider adapters. The comparison thus evaluates deployed model--harness combinations, including their differences in tool and image handling.

\paragraph{Execution.}
The benchmark allows 60 minutes of agent execution, excluding preparation, startup, and export. Available final states are evaluated after normal termination, errors, and timeouts. Separate source-level runs do not all share this limit: the larger no-reference PartNet runs did not record a timeout setting. Execution status and geometric success are treated separately.

\paragraph{Task Instructions.}
\label{app:prompts}
All systems receive the following task text, with the reference lines substituted according to the condition. Harness defaults are retained, while project instructions and stored memories are disabled. The task text is shared across systems; their internal system prompts need not be identical.
\begin{quote}\small
\ttfamily
Assemble the supplied parts into a coherent object.\\
\normalfont\emph{[Reference-specific lines below.]}\ttfamily\\
Inspect connections from multiple camera views using capture\_scene, correct gaps, orientation and obvious interpenetration, and report uncertainties. Do not claim computed accuracy or physical stability without evidence.\\
Before finishing, check whether all supplied parts have been incorporated into the assembly.
\end{quote}
The reference-specific lines are:
\begin{itemize}
\item \textbf{No reference:} ``No manual or reference image is supplied. Infer the assembly from the part geometry alone.''
\item \textbf{Assembled-object image:} ``One reference image of the finished assembly is attached. Assemble the parts so that the result matches that image.''
\item \textbf{Manual:} ``The attached images are the pages of the assembly manual, in order. Read them in order and follow them.''
\end{itemize}
An additional execution instruction restricts interaction to the supplied environment tools, prohibits access to source geometry, ground truth, and other experiments, and prohibits resetting the scene or spawning other agents. Supplied references may be reread.

\paragraph{Resource Measurements.}
Resource columns report arithmetic per-evaluation means, whereas assembly quality uses equal source weights. Elapsed run time includes intervals between calls and is not a measure of internal reasoning time. Claude Code reports cost directly; Codex cost is estimated from the recorded price schedule. These are usage estimates rather than subscription charges. Input-token accounting includes cache reads and, for Claude, cache creation; output tokens are added once. Records cover all 100 evaluations for Astra, Sol, Terra, Fable, and Sonnet, but only 88 for Opus, 78 for DeepSeek, and 34 for Qwen. Means exclude missing records and may therefore reflect outcome-dependent coverage.

\subsection{Geometric Evaluation}
Each part contributes 1,000 farthest-point samples selected from 4,096 area-weighted surface candidates. Evaluation normalizes the largest part's vertex-PCA bounding-box diagonal to one, independently of the initialization scale. For point sets $X$ and $Y$, squared bidirectional Chamfer distance is
\begin{equation}
d_{\rm CD}(X,Y)=\frac1{|X|}\sum_{x\in X}\min_{y\in Y}\|x-y\|^2+\frac1{|Y|}\sum_{y\in Y}\min_{x\in X}\|y-x\|^2.
\end{equation}
SCD applies this distance to the assembled point sets. A single rigid registration aligns the complete prediction to the target while preserving relative part placement. We select the lowest SCD found from 24 proper PCA initializations and same-ID part-pose initializations, each refined by at most 100 symmetric ICP (iterative closest point) iterations with improvement tolerance $10^{-9}$.

Geometrically equivalent parts form connected components of pairwise proper-rigid shape matches with CD $\leq10^{-4}$. Hungarian assignment within each group minimizes the sum of part CDs. Given matched errors $e_i$,
\begin{equation}
\mathrm{PA}=\frac1n\sum_{i=1}^n\mathbf{1}[e_i\leq0.01],\qquad
\mathrm{SR}=\mathbf{1}[\max_i e_i\leq0.01].
\end{equation}
No part is independently repositioned to improve its scored pose. Registration is approximate, and equivalence uses the transitive closure of pairwise matches. These definitions are fixed across systems. For source means $\bar s$, aggregation is
\begin{equation}
\mathrm{Overall}=\tfrac14\left[\tfrac12(\bar s_{\rm PN,none}+\bar s_{\rm PN,image})+\bar s_{\rm IKEA}+\bar s_{\rm AB}+\bar s_{\rm FB}\right].
\end{equation}
PA and benchmark quality curves use the same weighting. All benchmark tasks have scored final states; under the benchmark rule, a missing outcome would receive zero PA and SR, while SCD requires finite geometry.

\paragraph{Source-Level Comparisons.}
The PartNet comparison includes joint-guided assembly in Joint-PA~\citep{li2024joint} and co-creation-space assembly in CCS~\citep{zhang2024ccs}.
The Fantastic Breaks evaluation samples 5,000 area-weighted surface points across parts, with at least 20 per part. Its scale is $\max(1,e_{\max})$, where $e_{\max}$ is the largest source per-part axis-aligned extent. A shared transform aligns the largest sampled part to its target, with that anchor included in the average and fixed part identities. For wrapped intrinsic-XYZ Euler-angle differences $\Delta\phi_i$ and sampled-centroid differences $\Delta c_i$,
\begin{equation}
\mathrm{RE}=\frac1n\sum_{i=1}^{n}\sqrt{\frac{\|\Delta\phi_i\|^2}{3}},\qquad
\mathrm{TE}=\frac1n\sum_{i=1}^{n}\sqrt{\frac{\|\Delta c_i\|^2}{3}}.
\end{equation}
RE is in degrees and differs from geodesic rotation error. PA uses strict part CD $<0.01$; shape CD uses global nearest neighbors, averaged within parts and then equally across parts. TE and CD are displayed at $\times100$ and $\times1000$. Published Jigsaw, PuzzleFusion++ (PF++)~\citep{wang2025puzzle}, and GARF results follow GARF Table 3; RPF, TORA-CKA~\citep{lee2026tora}, and SARe-Gen follow SARe v2 Table 3. Their original evaluation implementations are retained. Our anchor alignment and sampled inputs do not constitute a matched reproduction of those published runs; the four-way refinement experiment instead uses matched inputs throughout.

IKEA Chair/Table baselines follow Manual-PA Table 1. Full and Bench/Chair/Desk comparisons, including TwoByTwo~\citep{qi2025twobytwo}, follow AssemLM v2 Tables 5--6 and Table 2. The alternative $\mathrm{SR}^{a}$ uses SCD $<0.02$ before the display multiplier, with each method's normalization. It measures whole-shape proximity rather than requiring every part to be correct.

\FloatBarrier
\section{Additional Quantitative Results}
\label{app:complete_results}
\subsection{Uncertainty and Reference Effects}
Tables~\ref{tab:analysis_overview}--\ref{tab:analysis_reference} report benchmark quality, paired system differences, and image-reference effects. Percentile bootstrap intervals use 10,000 draws with a fixed random seed, resampling objects within each source. Both PartNet conditions remain paired, and system comparisons use identical resampled objects and the benchmark weights. These pointwise intervals describe object-sampling variation within the recorded runs, not repeated-run reliability. An all-zero interval does not establish zero population success probability.

Pairwise comparisons use two-sided randomization tests with 100,000 draws and the add-one correction. System labels are exchanged within objects, jointly for both PartNet conditions. Holm correction covers 56 tests: 28 system pairs and two metrics. Confidence intervals remain pointwise. Astra's descriptive advantage over Fable is not significant under this family-wise correction for either metric.
\begin{table}[tbp]
\centering\footnotesize
\caption{Source-weighted benchmark quality and pointwise paired-object bootstrap intervals (percent). All 100 tasks per system contribute, including partial assemblies after errors or timeouts.}
\label{tab:analysis_overview}
\begin{NiceTabular*}{\linewidth}{@{\extracolsep{\fill}}lrrrrr@{}}
\toprule
System & SR & 95\% CI & PA & 95\% CI & Errors/timeouts \\
\midrule
GPT-6 Astra & 59.40 & [50.0, 68.1] & 80.90 & [75.0, 86.3] & 0 \\
Claude Fable & 50.00 & [41.2, 58.8] & 74.10 & [67.8, 80.0] & 0 \\
Claude Opus & 44.40 & [34.4, 54.4] & 63.10 & [55.1, 70.8] & 12 \\
GPT-5.6 Sol & 11.20 & [5.0, 17.5] & 30.60 & [23.5, 37.7] & 0 \\
Claude Sonnet & 7.500 & [2.5, 12.5] & 14.40 & [9.0, 20.0] & 0 \\
GPT-5.6 Terra & 7.500 & [2.5, 12.5] & 11.20 & [6.2, 16.2] & 0 \\
Qwen Max & 11.90 & [5.6, 18.1] & 15.90 & [9.8, 22.3] & 79 \\
DeepSeek Flash & 0.000 & [0.0, 0.0] & 0.000 & [0.0, 0.0] & 51 \\
\bottomrule
\end{NiceTabular*}
\end{table}

\begin{table}[tbp]
\centering\footnotesize
\caption{All system differences (A minus B, percentage points), with pointwise paired-object 95\% intervals. Randomization-test $p$ values use Holm correction across all 56 system-pair/metric comparisons; intervals remain pointwise.}
\label{tab:analysis_paired}
\begin{NiceTabular*}{\linewidth}{@{\extracolsep{\fill}}llrrrr@{}}
\toprule
A & B & $\Delta$SR [95\% CI] & $p_{\rm Holm}$ & $\Delta$PA [95\% CI] & $p_{\rm Holm}$ \\
\midrule
Astra & Fable & +9.4 [0.6, 18.1] & 0.581 & +6.8 [1.3, 12.3] & 0.263 \\
Astra & Opus & +15.0 [5.6, 24.4] & 0.064 & +17.8 [10.5, 25.4] & $<0.001$ \\
Astra & Sol & +48.1 [38.1, 58.1] & $<0.001$ & +50.3 [41.8, 58.7] & $<0.001$ \\
Astra & Sonnet & +51.9 [41.9, 61.9] & $<0.001$ & +66.5 [59.2, 73.6] & $<0.001$ \\
Astra & Terra & +51.9 [40.6, 62.5] & $<0.001$ & +69.7 [61.5, 77.5] & $<0.001$ \\
Astra & Qwen & +47.5 [37.5, 57.5] & $<0.001$ & +65.0 [57.4, 72.5] & $<0.001$ \\
Astra & DeepSeek & +59.4 [50.0, 68.1] & $<0.001$ & +80.9 [75.0, 86.3] & $<0.001$ \\
Fable & Opus & +5.6 [-2.5, 13.8] & 1.000 & +11.0 [4.5, 17.8] & 0.021 \\
Fable & Sol & +38.8 [28.7, 48.1] & $<0.001$ & +43.5 [35.3, 51.5] & $<0.001$ \\
Fable & Sonnet & +42.5 [33.1, 51.9] & $<0.001$ & +59.7 [52.1, 67.1] & $<0.001$ \\
Fable & Terra & +42.5 [32.5, 52.5] & $<0.001$ & +62.9 [55.3, 70.2] & $<0.001$ \\
Fable & Qwen & +38.1 [28.7, 48.1] & $<0.001$ & +58.2 [50.4, 65.7] & $<0.001$ \\
Fable & DeepSeek & +50.0 [41.2, 58.8] & $<0.001$ & +74.1 [67.8, 80.0] & $<0.001$ \\
Opus & Sol & +33.1 [23.1, 43.8] & $<0.001$ & +32.5 [23.6, 41.2] & $<0.001$ \\
Opus & Sonnet & +36.9 [25.6, 48.1] & $<0.001$ & +48.7 [39.2, 57.9] & $<0.001$ \\
Opus & Terra & +36.9 [25.6, 48.1] & $<0.001$ & +51.9 [43.2, 60.5] & $<0.001$ \\
Opus & Qwen & +32.5 [22.5, 42.5] & $<0.001$ & +47.2 [38.7, 55.6] & $<0.001$ \\
Opus & DeepSeek & +44.4 [34.4, 54.4] & $<0.001$ & +63.1 [55.1, 70.8] & $<0.001$ \\
Sol & Sonnet & +3.8 [-1.2, 10.0] & 1.000 & +16.2 [8.4, 24.3] & 0.003 \\
Sol & Terra & +3.8 [-3.8, 11.2] & 1.000 & +19.4 [12.3, 26.6] & $<0.001$ \\
Sol & Qwen & -0.6 [-6.9, 5.6] & 1.000 & +14.7 [8.1, 21.6] & 0.002 \\
Sol & DeepSeek & +11.2 [5.0, 17.5] & 0.064 & +30.6 [23.5, 37.7] & $<0.001$ \\
Sonnet & Terra & +0.0 [-6.2, 7.5] & 1.000 & +3.2 [-3.8, 10.2] & 1.000 \\
Sonnet & Qwen & -4.4 [-10.6, 1.9] & 1.000 & -1.4 [-8.7, 5.6] & 1.000 \\
Sonnet & DeepSeek & +7.5 [2.5, 12.5] & 0.407 & +14.4 [9.0, 20.0] & $<0.001$ \\
Terra & Qwen & -4.4 [-11.2, 1.9] & 1.000 & -4.7 [-11.4, 1.7] & 1.000 \\
Terra & DeepSeek & +7.5 [2.5, 12.5] & 0.407 & +11.2 [6.2, 16.2] & 0.002 \\
Qwen & DeepSeek & +11.9 [5.6, 18.1] & 0.035 & +15.9 [9.8, 22.3] & $<0.001$ \\
\bottomrule
\end{NiceTabular*}
\end{table}

\begin{table}[tbp]
\centering\footnotesize
\caption{Paired image-conditioned minus no-reference results on the same 20 PartNet objects. Up/down counts exclude ties; SR counts are discordant successes. PA intervals are pointwise percentile bootstrap intervals.}
\label{tab:analysis_reference}
\begin{NiceTabular*}{\linewidth}{@{\extracolsep{\fill}}lrrrr@{}}
\toprule
System & $\Delta$PA (pp) & 95\% CI & PA up/down & SR up/down \\
\midrule
GPT-6 Astra & +20.6 & [9.5, 32.6] & 12/2 & 5/0 \\
Claude Fable & +19.3 & [5.5, 34.3] & 13/5 & 4/0 \\
Claude Opus & +15.6 & [2.0, 30.0] & 13/4 & 3/0 \\
GPT-5.6 Sol & +2.7 & [-5.1, 11.2] & 6/4 & 0/0 \\
Claude Sonnet & -3.9 & [-11.7, 0.0] & 0/1 & 0/0 \\
GPT-5.6 Terra & +0.0 & [0.0, 0.0] & 0/0 & 0/0 \\
Qwen Max & -12.5 & [-24.1, -4.0] & 0/8 & 0/1 \\
DeepSeek Flash & +0.0 & [0.0, 0.0] & 0/0 & 0/0 \\
\bottomrule
\end{NiceTabular*}
\end{table}

\subsection{Geometry, Complexity, and Metric Sensitivity}
\label{app:part_scale}
\paragraph{Relative Part Size.}
Table~\ref{tab:part_scale} measures each object's largest-to-smallest part PCA bounding-box diagonal, using unique mesh vertices. Whole-object normalization preserves this ratio. AssemblyBench has a higher median than the furniture sets, although the difference is not uniform across categories: PartNet Storage has a 90th percentile of 17.68, exceeding AssemblyBench's 15.73.
\begin{table}[tbp]
\centering\footnotesize
\caption{Within-object part-size ratios on the evaluated source-level sets. Each ratio divides the largest part PCA bounding-box diagonal by the smallest. PartNet assigns equal weight to objects across all three categories.}
\label{tab:part_scale}
\begin{tabular}{lrrrrr}
\toprule
Dataset & $N$ & Median & Interquartile range & 90th percentile & Ratio $>10$ (\%) \\
\midrule
AssemblyBench & 279 & 4.39 & 2.29--8.90 & 15.73 & 20.1 \\
IKEA-Manual & 102 & 2.20 & 1.67--3.28 & 4.52 & 3.9 \\
PartNet & 1472 & 1.79 & 1.00--3.22 & 8.32 & 8.6 \\
\bottomrule
\end{tabular}
\end{table}

\paragraph{Part Count.}
Table~\ref{tab:source_complexity} reports PartNet results in fixed bands of 2--5, 6--10, and 11--20 parts. Category and reference conditions remain separate. Table-category PA and SR decline across bands, whereas Chair/no-reference is not monotonic. Storage's lowest band contains only one object. The part-size and part-count comparisons are descriptive: geometry, category, and complexity vary together, preventing causal attribution to any one factor.
\begin{table}[tbp]
\centering\footnotesize
\caption{Part-count bands within complete PartNet category--reference conditions. Each cell shows sample count and mean PA/SR in percent. Bands keep objects with equal part counts together; category and reference conditions are not pooled.}
\label{tab:source_complexity}
\begin{tabular}{lrrr}
\toprule
Condition & 2--5 parts & 6--10 parts & 11--20 parts \\
\midrule
Chair / None & 40: 51.50/25.00 & 358: 61.90/28.50 & 393: 50.40/17.00 \\
Table / None & 136: 85.80/79.40 & 237: 73.50/49.40 & 160: 55.90/22.50 \\
Table / Image & 136: 93.70/86.80 & 237: 85.30/66.70 & 160: 78.30/39.40 \\
Storage / None & 1: 60.00/0.000 & 42: 58.50/28.60 & 105: 46.60/6.700 \\
Storage / Image & 1: 100.0/100.0 & 42: 69.70/33.30 & 105: 61.20/14.30 \\
\bottomrule
\end{tabular}
\end{table}

\paragraph{SCD Concentration.}
\label{app:scd_tail}
Separated parts can remain far apart in free space even after whole-object registration, producing large squared distances. Table~\ref{tab:scd_tail} reports the contribution of the largest $\lceil0.05N\rceil$ SCD values in the five analyzed PartNet conditions. All objects remain in the reported means. For Storage/image, eight objects contribute 99.32\% of summed SCD; the median is 0.53 despite a mean of 184.80. Six of these eight retain every initial part pose, and the other two retain most of them. Normal execution termination does not exclude geometric failure. This source of error differs from bounded translation predictors such as DGL, whose released pose head uses $\tanh$; input normalization alone does not impose an output bound.
\begin{table}[tbp]
\centering\footnotesize
\caption{Concentration of PartNet SCD for Astra. The largest $k=\lceil0.05N\rceil$ values define the upper tail; its contribution is the fraction of summed SCD. Mean and median use all objects; the last column excludes the upper tail only as a diagnostic. SCD uses the same scale as Table~\ref{tab:partnet}.}
\label{tab:scd_tail}
\begin{tabular}{lrrrrrr}
\toprule
Setting & $N$ & Mean & Median & $k$ & Contribution (\%) & Remaining mean \\
\midrule
Chair / NR & 791 & 69.90 & 2.88 & 40 & 94.62 & 3.96 \\
Table / NR & 533 & 46.98 & 1.21 & 27 & 94.53 & 2.71 \\
Storage / NR & 148 & 141.60 & 0.58 & 8 & 98.87 & 1.70 \\
Table / IR & 533 & 20.95 & 0.49 & 27 & 96.02 & 0.88 \\
Storage / IR & 148 & 184.80 & 0.53 & 8 & 99.32 & 1.33 \\
\bottomrule
\end{tabular}
\end{table}

\begin{figure}[tbp]
\centering
\includegraphics[width=\linewidth]{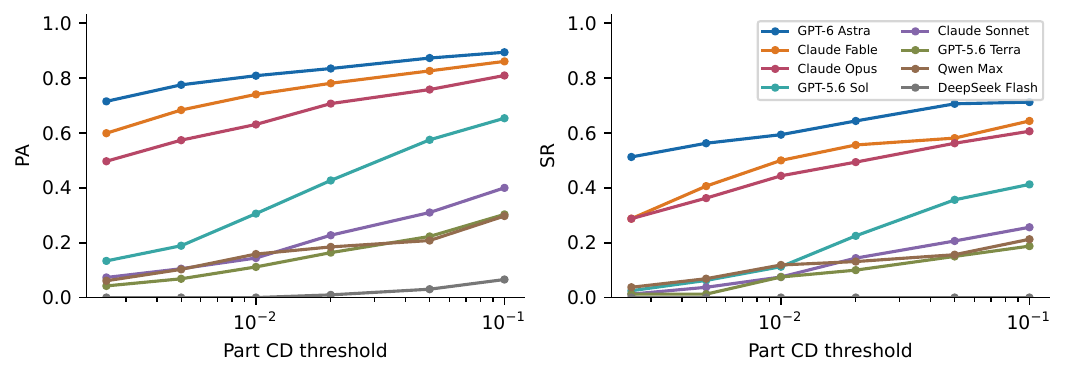}
\caption{Correctness-threshold sensitivity for all eight systems. PA/SR use equal source weights; saved registration and Hungarian correspondence remain fixed. The input-shape equivalence threshold remains fixed.}
\label{fig:thresholds}
\end{figure}

\paragraph{Category Results and Thresholds.}
Table~\ref{tab:ikea_categories} provides the separate IKEA Chair and Table results underlying the equally weighted summary. Figure~\ref{fig:thresholds} varies the correctness threshold with registration and correspondence fixed. This isolates cutoff sensitivity from changes in alignment or equivalence grouping.
\begin{table}[tbp]

\caption{IKEA-Manual results by category, before the equal-category averaging in Table~\ref{tab:ikea}. $\dagger$ denotes Image-PA retrained on diagrams.}
\label{tab:ikea_categories}
\centering
\footnotesize
\setlength{\tabcolsep}{3pt}
\begin{NiceTabular*}{\linewidth}{@{\extracolsep{\fill}}ll*{6}{r}@{}}
\CodeBefore
\rowcolor{gray!12}{6}
\Body
\toprule
\Block{2-1}{Method} & \Block{2-1}{Condition} & \Block{1-2}{SCD $\downarrow$} & & \Block{1-2}{PA $\uparrow$} & & \Block{1-2}{SR $\uparrow$} & \\
\cmidrule(lr){3-4}\cmidrule(lr){5-6}\cmidrule(lr){7-8}
 & & Chair & Table & Chair & Table & Chair & Table \\
\midrule
\venue{3DHPA}{CVPR'24} & -- & 34.3 & 37.8 & 1.914 & 4.027 & 0.000 & 0.000 \\
\venue{Image-PA}{ECCV\textquotesingle20}$^\dagger$ & Diagram & 17.3 & 14.7 & 19.07 & 36.74 & 0.000 & 10.53 \\
\venue{Manual-PA}{ICCV'25} & Manual & 11.4 & 4.8 & 42.51 & 49.72 & 3.509 & 15.79 \\
GPT-6 Astra + Codex & Manual & \textbf{1.2} & \textbf{0.4} & \textbf{91.07} & \textbf{95.47} & \textbf{71.93} & \textbf{78.95} \\
\bottomrule
\end{NiceTabular*}
\end{table}

\FloatBarrier
\section{Interaction and Failure Analysis}
\label{app:behavior}
\label{app:failures}
\subsection{Interaction Dynamics}
\label{app:interaction_dynamics}
Each evaluation is divided into ten equal-width bins from its first to last environment call. Activity shares are computed within each active evaluation-bin and then averaged with benchmark source weights. Inspect includes object and state queries; Observe includes camera moves and captures; Manipulate includes pose edits. Grouping and episode-control calls remain in the denominator. Bins without calls are omitted and available weights renormalized. Thus, evaluations with more calls do not automatically dominate the activity curves.

For each bin, manipulation magnitude is first summarized by its median within an evaluation and then by the weighted median across evaluations. Translation measures part-centroid displacement normalized by the largest-part diagonal; rotation measures angular displacement. Operations below $0.5^\circ$ enter translation summaries, and other operations enter rotation summaries. Group edits contribute the maximum magnitude among affected parts. Missing actions are not assigned zero magnitude. Table~\ref{tab:interaction_stages} applies the same procedure to the first and last thirds of each trajectory. Smaller late-stage motions occur in both successful and failed runs, so they do not by themselves establish successful refinement.
\begin{table}[tbp]
\centering\footnotesize
\caption{Early-to-late interaction changes. Each entry is a source-weighted median of per-evaluation action medians within the first or last third of the recorded interaction interval. Translation is normalized by the largest-part diagonal; rotation is in degrees. Revisit is the median fraction of edits affecting a previously edited part.}
\label{tab:interaction_stages}
\begin{tabular}{lrrrrr}
\toprule
System & Trans. early & Late & Rot. early & Late & Revisit (\%) \\
\midrule
Astra & 0.573 & 0.051 & 81.7 & 44.0 & 81.7 \\
Fable & 1.072 & 0.053 & 72.5 & 19.0 & 87.5 \\
Opus & 1.529 & 0.332 & 73.8 & 45.0 & 88.1 \\
Sol & 0.827 & 0.273 & 90.0 & 85.0 & 88.9 \\
Sonnet & 1.984 & 0.916 & 90.0 & 90.0 & 90.0 \\
Terra & 0.386 & 0.390 & 90.0 & 90.0 & 75.0 \\
Qwen & 0.819 & 0.806 & 90.0 & 90.0 & 87.5 \\
DeepSeek & 0.532 & 0.143 & 90.0 & 90.0 & 94.3 \\
\bottomrule
\end{tabular}
\end{table}

\begin{figure}[tbp]
\centering
\includegraphics[width=\linewidth]{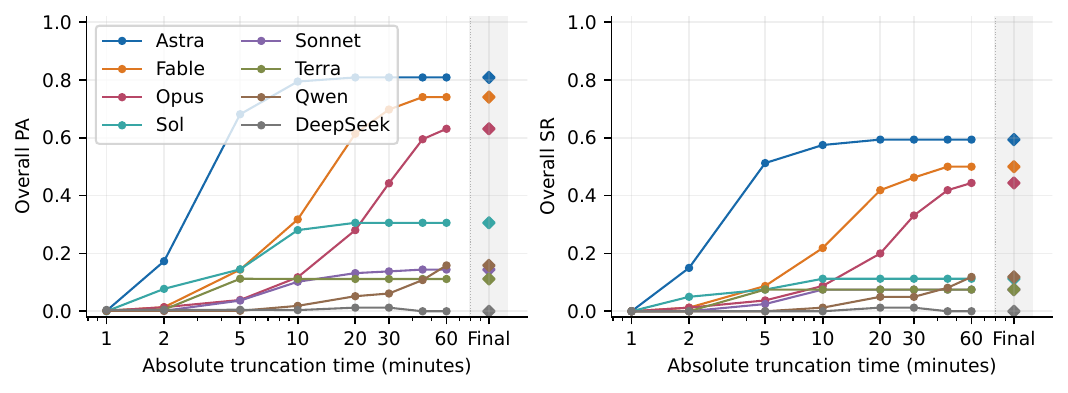}
\caption{Recorded-state truncation at absolute budgets. Earlier termination retains the final state. Diamonds in the shaded Final column show untruncated outcomes, not an additional time budget; five episodes have pose changes after 60 minutes of end-to-end time. Terminal values agree with the result tables. These are retrospective curves, not new budget-conditioned agent runs.}
\label{fig:budget_all}
\end{figure}

\paragraph{Quality and Time Budgets.}
For quality curves, we score the state after the last completed operation at each normalized checkpoint or absolute time budget. The initial state applies before the first operation and the final state persists after termination; scores are not interpolated between checkpoints. PA and SR use all evaluations with the benchmark weights. Figure~\ref{fig:budget_all} reports retrospective absolute-time truncation for all eight systems. Five episodes have edits after 60 minutes of end-to-end time, which includes startup outside the agent-execution budget; their untruncated outcomes are shown separately. PA and SR nevertheless agree at 60 minutes and termination for all systems.

A drop in PA can reflect a changed global registration or interchangeable-part assignment. Among 239 decreases between sampled checkpoints, 96 change at least one assignment; 140 decreases persist with alignment fixed, and 137 persist with both alignment and matching fixed. Intermediate accuracy is therefore interpreted alongside the actual manipulation, as in the inspection example in Figure~\ref{fig:interaction}.

\paragraph{Completion Reports.}
Astra succeeds in 43 of 74 declared completions, and Fable in 39 of 88. Among unsuccessful completion claims, median incorrect-part fractions are 33.3\% and 42.9\%, respectively. Missing reports are excluded. These counts compare categorical declarations with geometric success; they do not measure probabilistic confidence calibration.

\subsection{Execution and Geometric Failures}
\label{app:failure_modes}
Execution and geometric outcomes are distinct. Astra and Fable terminate normally in all 100 benchmark tasks. Qwen has 65 timeouts and 14 execution failures; DeepSeek has 17 and 34, respectively. All 49 normally terminated DeepSeek runs fail geometrically. Unsupported-tool and stream-disconnection errors occur in both systems, sometimes within the same run. These observations establish failures of the evaluated model--harness combinations, without attributing every error to the model's geometric reasoning.

The cases in Figure~\ref{fig:failure_completion} distinguish logical grouping from assembly, numerical pose agreement from visible alignment, and a residual positioning error in an otherwise assembled object.

\paragraph{Geometric Classification.}
We apply the saved global alignment and equivalent-part matching before assigning the first applicable category: correct (CD $\leq0.01$); never moved; far displaced (centroid offset $>3$ largest-part diagonals); translation-dominated (centered CD $\leq0.01$ and offset $>0.1$); near-threshold (centered CD $\leq0.01$ and offset $\leq0.1$); orientation-dominated (centered CD $>0.01$ and offset $\leq0.1$); or mixed. Centering removes translation while preserving orientation and accommodates geometric symmetry through shape distance.

Table~\ref{tab:failure_counts} gives denominators and compares part-pooled with task-pooled translation-error fractions. Varying the offset cutoff over $0.05,0.1,0.2$ and centered-CD cutoff over $0.005,0.01,0.02$ yields translation-dominated fractions of 48.3--78.0\% for Astra and 50.5--86.1\% for Fable. Their default values are 67.4\% and 75.1\%. Positioning error is a useful diagnostic, but the majority classification is not invariant to the thresholds.
\begin{table}[tbp]
\centering\footnotesize
\caption{Geometric diagnostic denominators and translation-dominated error fractions. Part pooling weights incorrect parts equally; task pooling first averages within each failed task. Categories follow the ordered classifier.}
\label{tab:failure_counts}
\begin{NiceTabular*}{\linewidth}{@{\extracolsep{\fill}}lrrrr@{}}
\toprule
System & Failed tasks & Wrong parts & Part pooled (\%) & Task pooled (\%) \\
\midrule
GPT-6 Astra & 47 & 236 & 67.4 & 68.5 \\
Claude Fable & 57 & 309 & 75.1 & 75.3 \\
Claude Opus & 61 & 400 & 67.8 & 64.7 \\
GPT-5.6 Sol & 91 & 631 & 46.4 & 50.9 \\
Claude Sonnet & 94 & 740 & 41.4 & 42.2 \\
GPT-5.6 Terra & 94 & 757 & 34.9 & 38.6 \\
Qwen Max & 90 & 730 & 34.4 & 36.8 \\
DeepSeek Flash & 100 & 778 & 32.9 & 38.2 \\
\bottomrule
\end{NiceTabular*}
\end{table}

\subsection{Qualitative Comparisons and Manual Use}
\label{app:manual_following}
\begin{figure}[p]
\centering
\includegraphics[width=.85\linewidth]{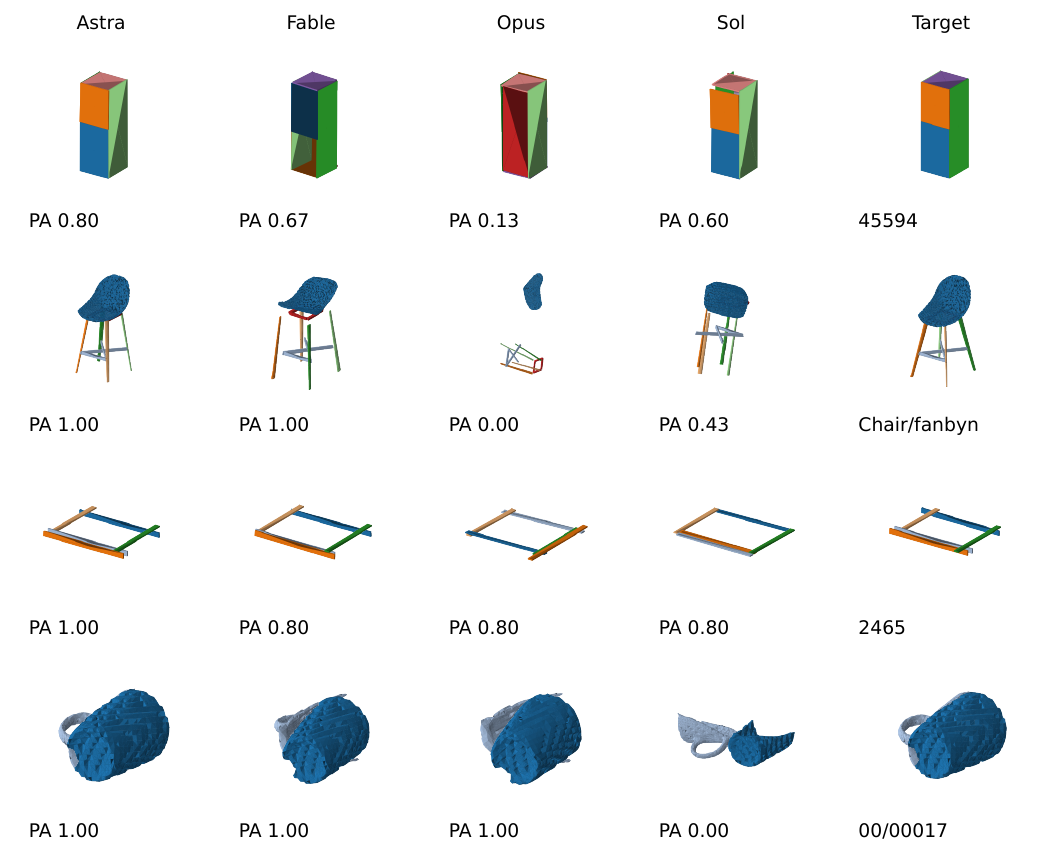}
\caption{Four common objects selected near Astra's median PA in each source, evaluated by Astra, Fable, Opus, and Sol. Columns retain a shared view direction and per-part colors; each view is fitted to its own geometry, so apparent size is not a shared physical scale. Numbers are PA under the common benchmark protocol. The other four systems appear in Figure~\ref{fig:system_comparison_other}.}
\label{fig:system_comparison}
\end{figure}

\begin{figure}[p]
\centering
\includegraphics[width=.85\linewidth]{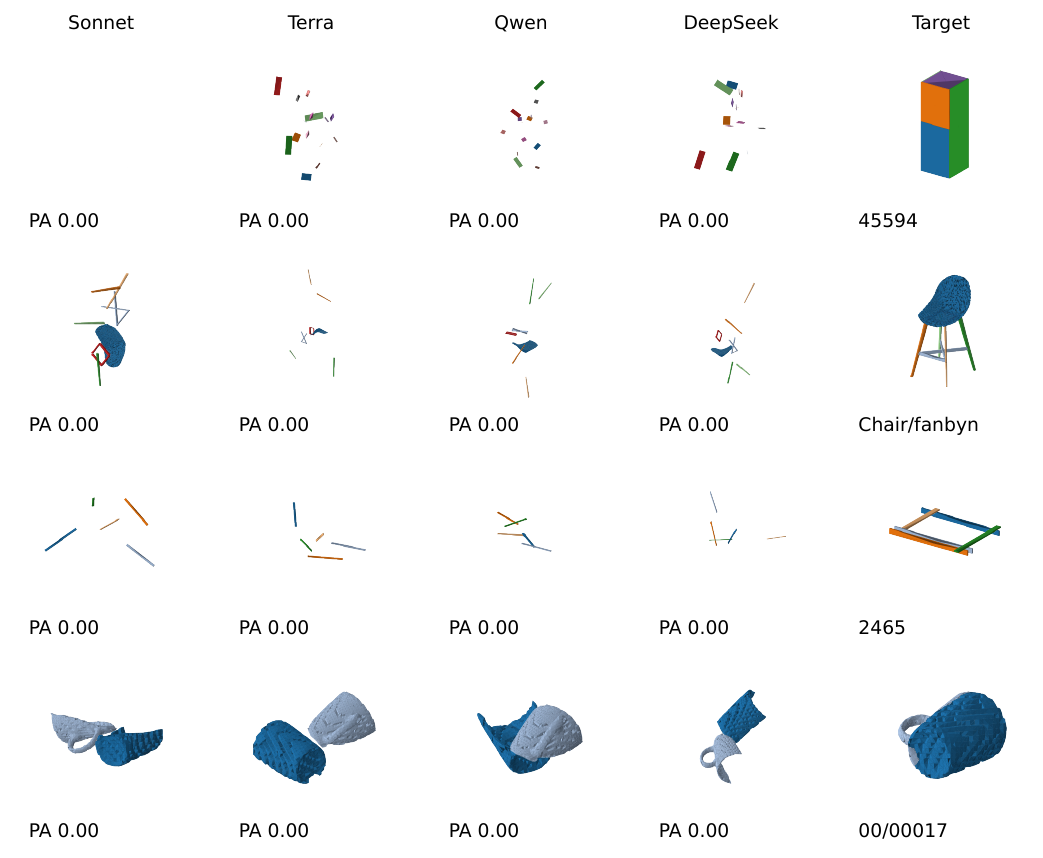}
\caption{The same four objects evaluated by Sonnet, Terra, Qwen, and DeepSeek, with the same rendering conventions as Figure~\ref{fig:system_comparison}.}
\label{fig:system_comparison_other}
\end{figure}

Figures~\ref{fig:system_comparison} and~\ref{fig:system_comparison_other} compare all eight systems on four common objects. Objects are selected near Astra's median PA within each source. Predictions are globally aligned to the targets; view directions and part colors are shared, but each rendering is fitted to its geometry.

Figure~\ref{fig:interaction} instead uses cases selected for visible intermediate manipulations. Its offline renderings preserve recorded part poses and camera directions with centered framing; they are not the original images observed by the agent. Reference images correspond to the supplied source materials.

In the IKEA bench example, the agent places the seat and both end frames before adding the stretcher, whereas the manual introduces the stretcher with the first end frame. In the industrial example, provisional placements are revised before and after inspecting the covered blade. These trajectories show correspondence to instructional relationships without reproducing every illustrated step. A different placement order alone does not establish failure to use a manual. These qualitative examples illustrate assembly behavior rather than estimate failure frequencies or instruction-following rates.

\FloatBarrier
\section{Combining Agents with Geometric Models}
\label{app:refinement}
\paragraph{Matched Comparison.}
All four configurations use the same 150 two-fragment Fantastic Breaks objects, fixed point samples, part identities, normalization, and anchor-aligned evaluator. Each object has 5,000 area-weighted surface samples with at least 20 per part. Sampling and inference use fixed random seeds. The largest sampled part is the anchor. Standard GARF is rerun on these inputs; its matched result in Table~\ref{tab:hybrid} is distinct from the published GARF result in Table~\ref{tab:fantastic}.

Both GARF configurations use the released GARF-mini checkpoint without training or tuning. Standard inference uses random initialization and a one-step initialization stage, followed by 20 refinement steps. Agent-initialized inference replaces the initialization stage with the agent poses and retains the same refinement schedule. One rigid transformation aligns the agent anchor to its target while preserving relative part poses; the anchor remains fixed during inference. No non-anchor target pose is supplied.

For GARF-to-Agent, the predicted poses initialize the original meshes with no preceding interaction history. Astra receives the original no-reference prompt and uses medium effort through Codex, without target geometry or the GARF trajectory. These runs follow the original source-level fracture setting without a wall-time cutoff. All 150 terminate normally.

\begin{figure}[!t]
\centering\includegraphics[width=.85\linewidth]{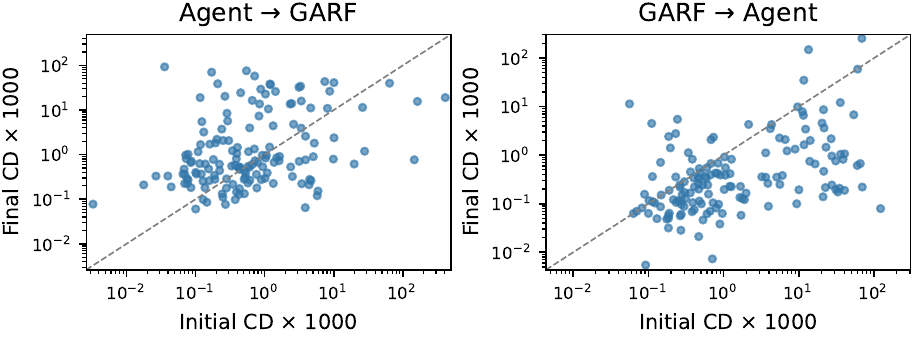}
\caption{Per-object changes in shape CD for the two composition orders. Each point is one of the same 150 objects; points below the diagonal improve. Both axes are logarithmic.}
\label{fig:refinement_paired}
\end{figure}

\paragraph{Paired Outcomes.}
Relative to standard GARF, agent initialization improves PA on ten objects and leaves it unchanged on 140. Relative to the unrefined agent, Agent-to-GARF improves PA on 15 objects, degrades it on 29, and leaves it unchanged on 106. Relative to standard GARF, GARF-to-Agent improves PA on 33 objects, degrades it on five, and leaves it unchanged on 112; shape CD improves on 125, worsens on 23, and is unchanged on two, using tolerance $10^{-8}$. Figure~\ref{fig:refinement_paired} shows both improvements and regressions. These single-run, two-fragment results do not establish seed-averaged superiority or transfer to multi-part industrial assemblies.

\FloatBarrier
\section{Robustness Experiments}
\label{app:robustness}
\paragraph{Conditions.}
Five fixed samples per benchmark block are selected using part-count strata, without selection on performance. PartNet uses two chairs, two tables, and one storage object, paired across reference conditions. The study contains 20 distinct objects and 25 baseline conditions. Each object receives two alternative layouts, one randomly removed part except on Fantastic Breaks, and one added part from a different object in the same source. PartNet conditions share the perturbations. Geometry, references, targets, and normalization remain fixed; inventory changes preserve unaffected parts' initial poses.

Selection, inventory changes, and alternative layouts use fixed random seeds. Each of the 95 new conditions is run once with medium-effort Astra, the original prompt, and a one-hour limit. All runs terminate normally. The unchanged prompt requests incorporation of all supplied parts and does not identify the perturbation. The 25 baselines are rescored using the same evaluator as the variants.

\begin{figure}[!t]
\centering
\begin{minipage}{.325\linewidth}\centering
\includegraphics[width=\linewidth]{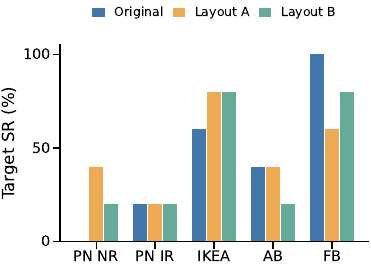}
\small (a) Initial layouts
\end{minipage}\hfill
\begin{minipage}{.325\linewidth}\centering
\includegraphics[width=\linewidth]{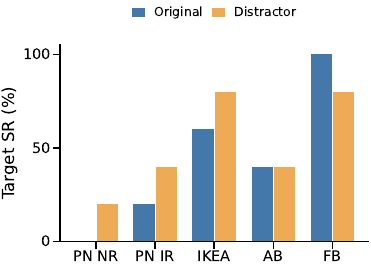}
\small (b) Distractor parts
\end{minipage}\hfill
\begin{minipage}{.325\linewidth}\centering
\includegraphics[width=\linewidth]{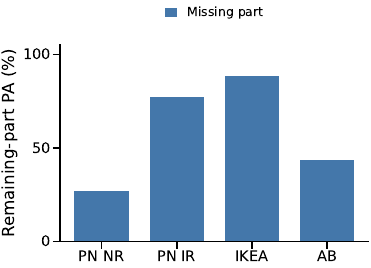}
\small (c) Missing parts
\end{minipage}
\caption{GPT-6 Astra under task variations on five fixed samples per benchmark block, with one run per condition. PartNet reference conditions share objects and perturbations. Missing-part evaluation measures only retained parts; Fantastic Breaks is excluded from this condition.}
\label{fig:robustness}
\end{figure}

\paragraph{Evaluation.}
Layout variants use benchmark PA and SR. Distractor evaluation excludes the added part from registration, correspondence, and scoring. Missing-part evaluation scores retained parts against their targets at the original scale. Retained-part completeness is not complete-object success, and retained-part PA has a different denominator from full-object PA. Figure~\ref{fig:robustness} summarizes the results; Table~\ref{tab:robustness_paired} preserves the object-level pairing.

\begin{table}[tbp]
\centering\footnotesize
\caption{Paired task-variation results. Each cell reports PA (\%) / SR (0 or 1); the missing-part column instead reports retained-part PA / retained-part completeness. A dash indicates an untested condition. Each condition is run once.}
\label{tab:robustness_paired}
\begin{tabular}{llrrrrr}
\toprule
Source & Object & Original & Layout A & Layout B & Distractor & Missing \\
\midrule
PartNet NR & 40074 & 81.80/0 & 100.0/1 & 81.80/0 & 72.70/0 & 30.00/0 \\
 & 2738 & 0.000/0 & 0.000/0 & 0.000/0 & 0.000/0 & 0.000/0 \\
 & 23814 & 18.20/0 & 18.20/0 & 18.20/0 & 18.20/0 & 10.00/0 \\
 & 22320 & 76.90/0 & 100.0/1 & 100.0/1 & 100.0/1 & 75.00/0 \\
 & 46475 & 54.50/0 & 54.50/0 & 54.50/0 & 27.30/0 & 20.00/0 \\
\midrule
PartNet IR & 40074 & 81.80/0 & 81.80/0 & 81.80/0 & 100.0/1 & 60.00/0 \\
 & 2738 & 76.90/0 & 53.80/0 & 69.20/0 & 76.90/0 & 75.00/0 \\
 & 23814 & 36.40/0 & 9.100/0 & 36.40/0 & 9.100/0 & 60.00/0 \\
 & 22320 & 100.0/1 & 100.0/1 & 100.0/1 & 100.0/1 & 91.70/0 \\
 & 46475 & 45.50/0 & 72.70/0 & 72.70/0 & 90.90/0 & 100.0/1 \\
\midrule
IKEA-Manual & Bench/applaro & 100.0/1 & 100.0/1 & 100.0/1 & 100.0/1 & 100.0/1 \\
 & Chair/voxlov & 66.70/0 & 66.70/0 & 66.70/0 & 66.70/0 & 80.00/0 \\
 & Chair/jokkmokk & 100.0/1 & 100.0/1 & 100.0/1 & 100.0/1 & 85.70/0 \\
 & Table/bjorkudden & 100.0/1 & 100.0/1 & 100.0/1 & 100.0/1 & 87.50/0 \\
 & Shelf/laiva & 57.90/0 & 100.0/1 & 100.0/1 & 100.0/1 & 88.90/0 \\
\midrule
AssemblyBench & 4451 & 100.0/1 & 100.0/1 & 100.0/1 & 100.0/1 & 0.000/0 \\
 & 2465 & 100.0/1 & 100.0/1 & 60.00/0 & 100.0/1 & 0.000/0 \\
 & 7780 & 60.00/0 & 60.00/0 & 80.00/0 & 40.00/0 & 100.0/1 \\
 & 4492 & 42.90/0 & 42.90/0 & 57.10/0 & 71.40/0 & 50.00/0 \\
 & 8243 & 46.20/0 & 46.20/0 & 46.20/0 & 46.20/0 & 66.70/0 \\
\midrule
Fantastic Breaks & 01/01007 & 100.0/1 & 100.0/1 & 100.0/1 & 100.0/1 & -- \\
 & 02/02002 & 100.0/1 & 100.0/1 & 100.0/1 & 100.0/1 & -- \\
 & 02/02007 & 100.0/1 & 0.000/0 & 0.000/0 & 0.000/0 & -- \\
 & 09/09028 & 100.0/1 & 100.0/1 & 100.0/1 & 100.0/1 & -- \\
 & 18/18002 & 100.0/1 & 50.00/0 & 100.0/1 & 100.0/1 & -- \\
\bottomrule
\end{tabular}
\end{table}

\paragraph{Responses to Inventory Changes.}
We examine actions, final geometry, and reports for all 20 missing-part and 25 distractor runs. Four IKEA and three AssemblyBench missing-part reports identify an absent component. A completion declaration can coexist with a reported omission. Under distractors, four IKEA and three AssemblyBench reports identify an uninstalled part, although several such parts were never moved. On Fantastic Breaks, two runs attempt to incorporate all three fragments, two try the extra fragment and then move it aside, and one reports leaving it unincorporated. These trajectories distinguish reported recognition, attempted use, and non-incorporation.

The unchanged all-parts instruction influences distractor handling. Without a reference, the intended inventory may also be underdetermined. The small paired sample and single run per condition support exploratory observations, not repeated-run reliability or robustness to larger numbers of missing or extra parts.

\end{document}